\pdfoutput=1

\documentclass[11pt]{article}

\usepackage[]{acl}

\usepackage{times}
\usepackage{latexsym}

\usepackage[T1]{fontenc}

\usepackage[utf8]{inputenc}

\usepackage{microtype}

\usepackage{inconsolata}

\usepackage{times}
\usepackage{latexsym}
\usepackage{commath}
\usepackage{multirow}
\usepackage{hyperref}
\usepackage{textcomp}
\usepackage{siunitx}
\usepackage{tabularx}
\usepackage{adjustbox}

\newcommand\tab[1][0.5cm]{\hspace*{#1}}
\usepackage{booktabs}
\usepackage{amsmath}
\usepackage{amsfonts}
\usepackage{graphicx}
\usepackage{subcaption}

\usepackage{cleveref}
\graphicspath{ {images/} }
\usepackage{color,soul}
\usepackage[T1]{fontenc}

\crefformat{section}{\S#2#1#3} 
\crefformat{subsection}{\S#2#1#3}
\crefformat{subsubsection}{\S#2#1#3}
\usepackage{float}

\usepackage{color, colortbl}
\definecolor{thepurple}{RGB}{103,78,167}
\definecolor{darkgreen}{RGB}{26, 125, 65}
\definecolor{myblue}{RGB}{0, 129, 251}

\usepackage[utf8]{inputenc}

\usepackage{microtype}

\usepackage{inconsolata}

\usepackage{afterpage}

\title{How Far Can We Extract Diverse Perspectives \\
from Large Language Models?}

\author{Shirley Anugrah Hayati\raisebox{3pt}{{\includegraphics[height=1em,width=1em]{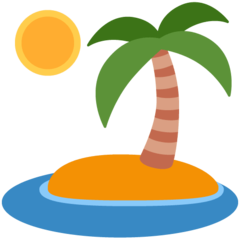}}} \tab Minhwa Lee\raisebox{3pt}{{\includegraphics[height=1em,width=1em]{emojis/island.png}}}
\tab
\textbf{Dheeraj Rajagopal}\raisebox{3pt}{{\includegraphics[height=1em,width=1em]{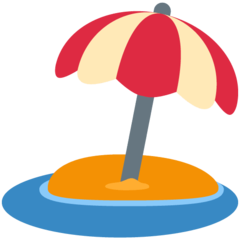}}} \tab \textbf{Dongyeop Kang}\raisebox{3pt}{{\includegraphics[height=1em,width=1em]{emojis/island.png}}} \\
  \raisebox{3pt}{{\includegraphics[height=1em,width=1em]{emojis/island.png}}}University of Minnesota \tab \raisebox{3pt}{{\includegraphics[height=1em,width=1em]{emojis/beach.png}}}Google DeepMind\\
  \texttt{\{hayat023, lee03533, dongyeop\}@umn.edu} \tab \texttt{rajagopald@google.com}
  }

\begin{document}
\maketitle
\begin{abstract}
Collecting diverse human opinions is costly and challenging. This leads to a recent trend in exploiting large language models (LLMs) for generating diverse data for potential scalable and efficient solutions. However, the extent to which LLMs can generate diverse perspectives on subjective topics is still unclear. 
In this study, we explore LLMs' capacity of generating diverse perspectives and rationales on subjective topics such as social norms and argumentative texts. We introduce the problem of extracting \textit{maximum diversity} from LLMs. Motivated by how humans form opinions based on values, we propose a criteria-based prompting technique to ground diverse opinions. 
To see how far we can extract diverse perspectives from LLMs, or called \textit{diversity coverage}, we employ a step-by-step recall prompting to generate more outputs from the model iteratively. Our methods, applied to various tasks, show that LLMs can indeed produce diverse opinions according to the degree of task subjectivity. We also find that LLM's performance of extracting maximum diversity is on par with human.\footnote{Our code and data are available at \href{https://github.com/minnesotanlp/diversity-extraction-from-llms}{https://github.com/minnesotanlp/diversity-extraction-from-llms}. The extracted opinions can be viewed on our project page here: 
\href{https://minnesotanlp.github.io/diversity-extraction-from-llms/}{https://minnesotanlp.github.io/diversity-extraction-from-llms/}}
\end{abstract}

\section{Introduction}

Using NLP for tasks that require social reasoning or involve human subjectivity like argumentation \cite{hidey-etal-2017-analyzing} or toxicity detection \cite{sap-etal-2019-risk} often calls for diverse perspectives. Instead of providing a single viewpoint, an ideal NLP model should accommodate various perspectives to avoid any bias towards a dominant one. Prior works emphasize the importance of modeling multiple viewpoints \cite{plank-2022-problem, nlperspectives-2022-perspectivist}. Some studies have addressed this challenge by gathering responses from multiple human annotators with diverse backgrounds \cite{rottger-etal-2022-two, santy-etal-2023-nlpositionality}. However, this approach is costly in time and resources. Recent advancements of LLMs have gained much interest from researchers to exploit their capability of creative generation for data augmentation with less cost and higher diversity \cite{cegin2023chatgpt, chung-etal-2023-increasing, bubeck2023sparks}. 

\begin{figure}
    \centering
     \includegraphics[width=\linewidth]{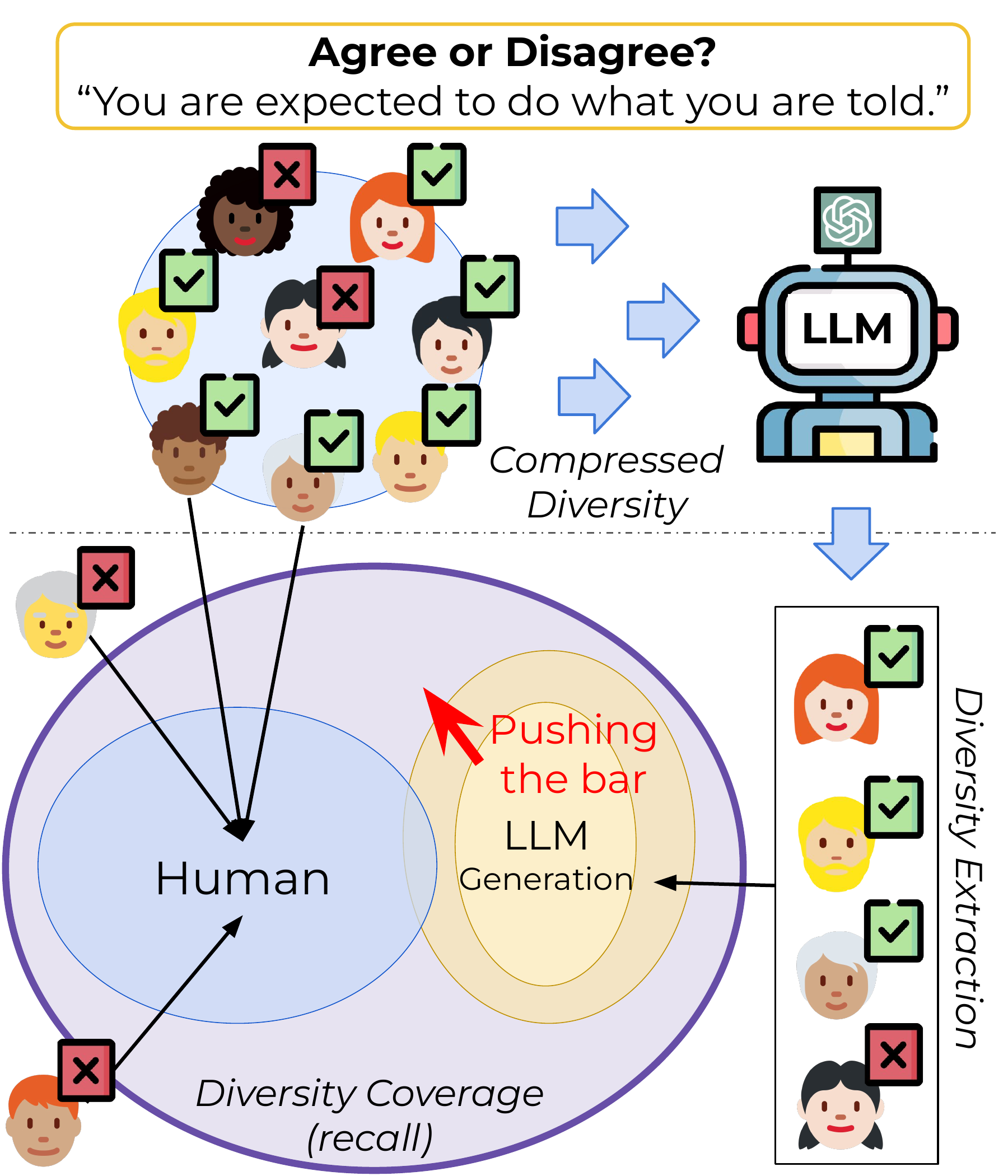} 
    \caption{LLMs are trained on texts written by different people who may have distinct perspectives. Our study examines whether LLMs can do ``reverse modeling'' of humans' perspectives from the training data and how much diversity coverage LLMs can generate. (A check mark = ``\textit{Agree}'' and a cross mark =``\textit{Disagree}'')
    }
    \label{fig:task_example}
    \vspace{-2mm}
\end{figure}

LLM is known as a compressed parametric knowledge (e.g., blurry JPEG) of the training corpus \cite{chiang2023}, and our work study how people’s pluralistic diverse opinions are compressed in the parameter and how far we can reversely extract them from an LLM. Figure \ref{fig:task_example} illustrates the significance of understanding the extent of diversity achievable by LLMs. During training, LLMs have access to various writings from humans with distinct values. Yet, can LLMs reflect this diversity when generating text, or do they tend to favor majority opinions? How do LLM-generated opinions overlap with human viewpoints? If the purple circle in Figure \ref{fig:task_example} hypothetically represents the maximum diversity achievable by humans, we aim to explore methods for LLMs to approach this diversity. 

\begin{figure}[t]
    \centering
    \includegraphics[width=\linewidth]{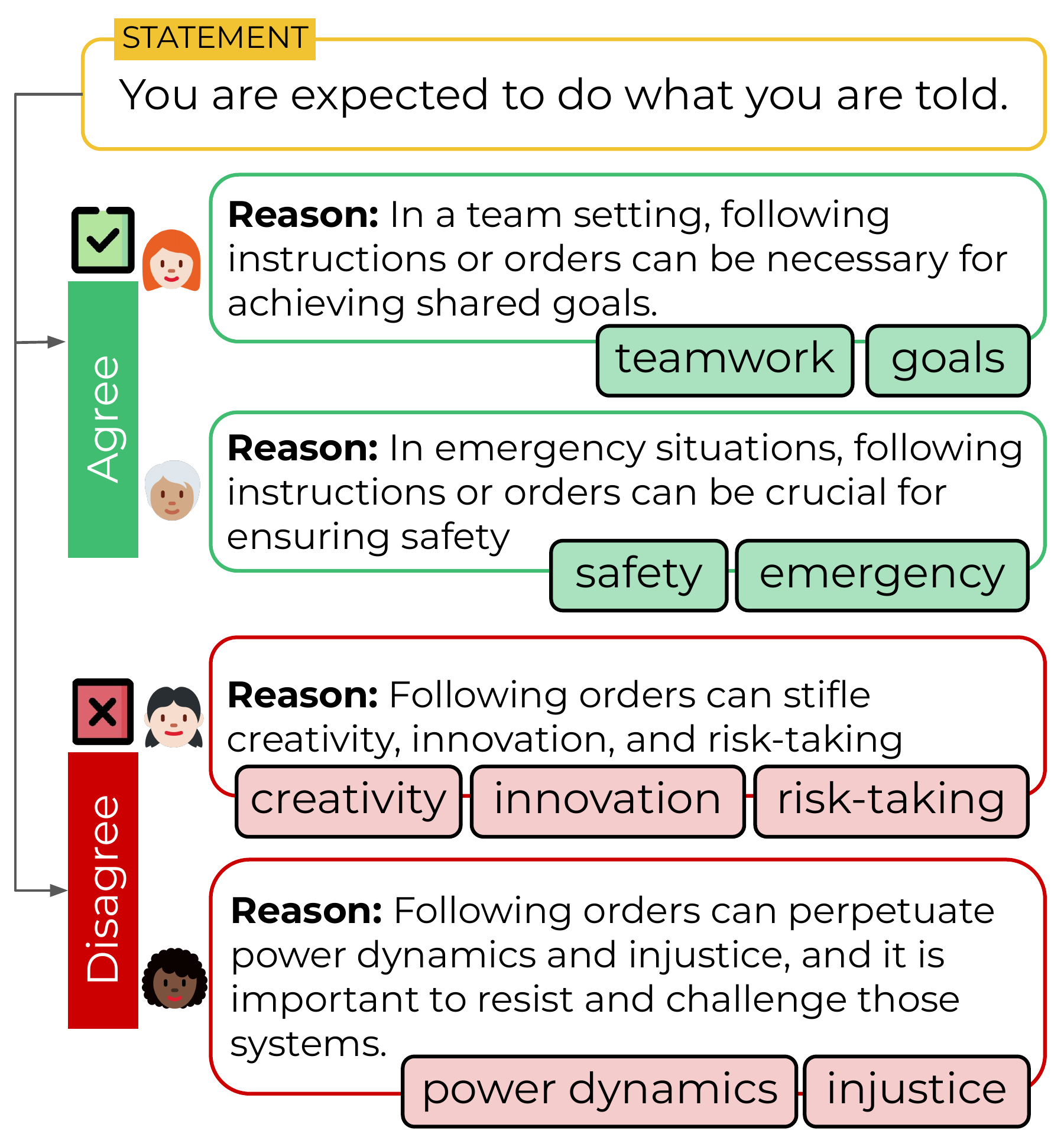} 
    \caption{People can have different opinions given a subjective statement. Given a statement, humans can agree or disagree with the statement with their own criteria (e.g., teamwork, risk-taking) in deciding their stances.
    }
    \label{fig:statement_and_opinions}
\end{figure}



In real-world scenarios, humans may take different stances on a subjective statement (Figure \ref{fig:statement_and_opinions}). For instance, those valuing \textit{teamwork} and \textit{goal achievements} may agree with the statement, while others valuing \textit{creativity} and \textit{innovation} could oppose it. Then the key question is: \textbf{\textit{How many and what diverse perspectives do LLMs model?}}

We introduce a novel problem called \textit{maximum diversity extraction} from LLMs. Formally, our task involves (1) asking LLMs to generate as many different stances as possible, (2) providing reasons for each stance, and (3) producing important criteria words that guide their reasoning process. We apply this diversity prompting across four subjective tasks: social norms, argumentation, hate speech labeling, and story continuation.



\paragraph{Contributions} 
First, we propose perspective diversity as a new focus for generative LLMs, distinct from lexical, syntactical, and semantic diversity mainly explored in previous studies. Through various experiments, we assess LLMs' capacity to achieve maximum perspective diversity. Second, we introduce criteria-based diversity prompting to extract and ground diverse perspectives from LLMs. Finally, we suggest a step-by-step recall approach to measure the extent of diversity coverage of LLMs, comparing the coverage between LLM-generated opinions with human-authored opinions.

Our study reveals a saturation point in the diversity that LLMs can achieve, depending on the subjectivity of a task. Also, LLMs generally produce more diverse opinions than an individual human, but two or more humans achieve greater diversity. Regarding the quality of LLM's generation, LLM is able to generate opinions which are semantically similar to human opinions. However, some frequent criteria words by LLM are different from what humans consider as important.


\section{Criteria-based Diversity Prompting}
\subsection{Motivation}
First, we present the motivation behind our approach. Imagine engaging in a debate with someone over a controversial topic. Effective debaters often employ overarching framing to shape their arguments persuasively and coherently. For example, framing arguments around ``power dynamics'' or ``creativity'' can effectively challenge the given statement as shown in the ``disagree'' examples in Figure \ref{fig:statement_and_opinions}. We refer to these framing keywords as \textit{criteria}. Opinions guided by these criteria could be more diverse as they are grounded in the combination of various criteria words.

\subsection{Step 1: Think of Your Criteria First before Making Opinions}
Our \textbf{criteria-guided prompting} is as follows:

``Given a \textit{statement}, generate a \textcolor{magenta}{Stance} and explain its \textcolor{purple}{Reasons} with a list of \textcolor{cyan}{Criteria} that affect a model's perspective''.

\paragraph{Task Definition}
The task is defined as choosing a binary stance and generating supporting reasons for a given subjective statement. The generated criteria by a model are a list of words or short phrases. The model's reasons include a free-form explanation of its stance (Table \ref{tab:prompting_example} and Table \ref{tab:recall_prompting}). 

\paragraph{Criteria-Based vs. Free-form}
We then compare model's diversity performance on two prompting settings: with criteria vs. without criteria (free-form). From our experiments (Section \ref{semantic_diversity_result}), we found that the criteria-based prompting method enables the model to generate important criteria for framing high-level decisions and providing well-grounded reasons. The criteria list can also be seen as reflecting the model's values. This approach follows human reasoning, where personal values often guide opinions and behavior \cite{rokeach_nature_1973, kesberg_relation_2018}.

\begin{table*}[t]
    \centering
    \small
    \begin{tabularx}{\textwidth}{l l l}
    \toprule
    \multicolumn{3}{c}{\textbf{Model Input} (one-shot example below)}
    \\\midrule
    \multicolumn{3}{l}{Statement: \textit{It's okay to have privacy}}\\
    \multicolumn{3}{l}{Tell me opinions about the statement as many as possible from different people with, ``Agree'' or ``Disagree,'' one-word or} \\
    \multicolumn{3}{l}{ one-phrase criteria that is important for their opinions, and explain how they have different opinions}\\
    Output: \\
    \multicolumn{3}{l}{\texttt{\{1:\{``\textcolor{magenta}{Stance}'' :``Agree'',}}\\
    \multicolumn{3}{l}{\tab \texttt{ ``\textcolor{cyan}{Criteria}'': [``personal boundaries'', ``autonomy''],}} \\
    \multicolumn{3}{l}{\tab \texttt{ ``\textcolor{purple}{Reason}'': ``Having privacy allows individuals to establish personal boundaries and 
 maintain.''}}\\
    \multicolumn{3}{l}{\tab \texttt{ their autonomy."\},}}\\
    \multicolumn{3}{l}{\texttt{ 2: \{``\textcolor{magenta}{Stance}'': ``Disagree",}}\\
    \multicolumn{3}{l}{\texttt{ \tab``\textcolor{cyan}{Criteria}'': [``transparency'', ``trust''],}}\\
    \multicolumn{3}{l}{\tab \texttt{ ``\textcolor{purple}{Reason}'': ``Lack of privacy can promote transparency and build trust in relationships.''}}\\ 
    ... \\
   \multicolumn{3}{l}{ \texttt{10: \{``\textcolor{magenta}{Stance}'': ``...", ``\textcolor{cyan}{Criteria}'': [...], ``\textcolor{purple}{Reason}'': `` ..."\}\}}} \\
   ...
   \\
    \multicolumn{3}{l}{Statement: \textit{You're expected to do what you are told}}\\
    \multicolumn{3}{l}{Tell me opinions about the statement as many as possible from different people with, ``Agree'' or ``Disagree,'' one-word or} \\
    \multicolumn{3}{l}{ one-phrase criteria that is important for their opinions, and explain how they have different opinions}\\
    Output: \\ \midrule
    \multicolumn{3}{c}{\textbf{Model Output}}
    \\ \midrule
     \multicolumn{3}{l}{\texttt{\{1:\{``\textcolor{magenta}{Stance}'' : ``Agree",}}\\
    \multicolumn{3}{l}{\tab \texttt{ ``\textcolor{cyan}{Criteria}'': [``teamwork'', ``goals''],}} \\
    \multicolumn{3}{l}{\tab \texttt{ ``\textcolor{purple}{Reason}'': ``In a team setting, following instructions or orders can be necessary for achieving}}\\
    \multicolumn{3}{l}{\tab \texttt{  shared goals.''\},}}\\
    ...\\
    \multicolumn{3}{l}{\texttt{10: \{``\textcolor{magenta}{Stance}'': ``...", ``\textcolor{cyan}{Criteria}'': [...], ``\textcolor{purple}{Reason}'': `` ..."\}}}\}
    \\ \bottomrule 
    \end{tabularx}
    \caption{Example prompt for criteria-based diversity prompting. Prompt for the free-form prompting is all the same except for not having the ``Criteria.'' Full prompts for free-form prompting and criteria prompting are shown in \ref{prompt_details}.
    \vspace{-4mm}
    }
    \label{tab:prompting_example}
\end{table*}

Motivated by recent advancements in few-shot learning that have enhanced model performance on challenging tasks, we utilize in-context prompting to explore the model's capacity to generate diverse opinions.\footnote{Please refer \ref{other_prompting} for details on how we choose this setting.} \cite{perez2021true, min-etal-2022-rethinking, min2022metaicl, lu-etal-2022-fantastically}. The output format is structured as a Python dictionary to be parsed for diversity evaluation. Each few-shot example contains ten opinions - five agreeing with the statement and five disagreeing. This setting does not influence the number of generated opinions and the content of each opinion, as we found cases when the model produces an imbalanced number of stances or opinions fewer than 10 (details in Appendix \ref{sec:appendix:unbalanced}). We also test the best-performing model with a zero-shot approach. 

\paragraph{Human Evaluation on Model-Generated Opinions}
To ensure the quality of model-generated opinions, we conduct human inspections to verify if each opinion entails its corresponding statement and stance. Over 99\% of opinions in randomly sampled opinion-statement pairs were found to align accurately. We also examine whether the generated criteria words entail the free-form reasons they support; 96\% of opinions in randomly sampled opinion-criteria pairs demonstrated this entailment. Further details about this process are described in the Appendix \ref{instruction_alignment} and \ref{eval_criteria_words}.

\subsection{Step 2: Step-by-Step Recall Prompting to Maximize Diversity Incrementally}
\begin{figure}[t]
    \centering
    \includegraphics[width=0.9\linewidth]{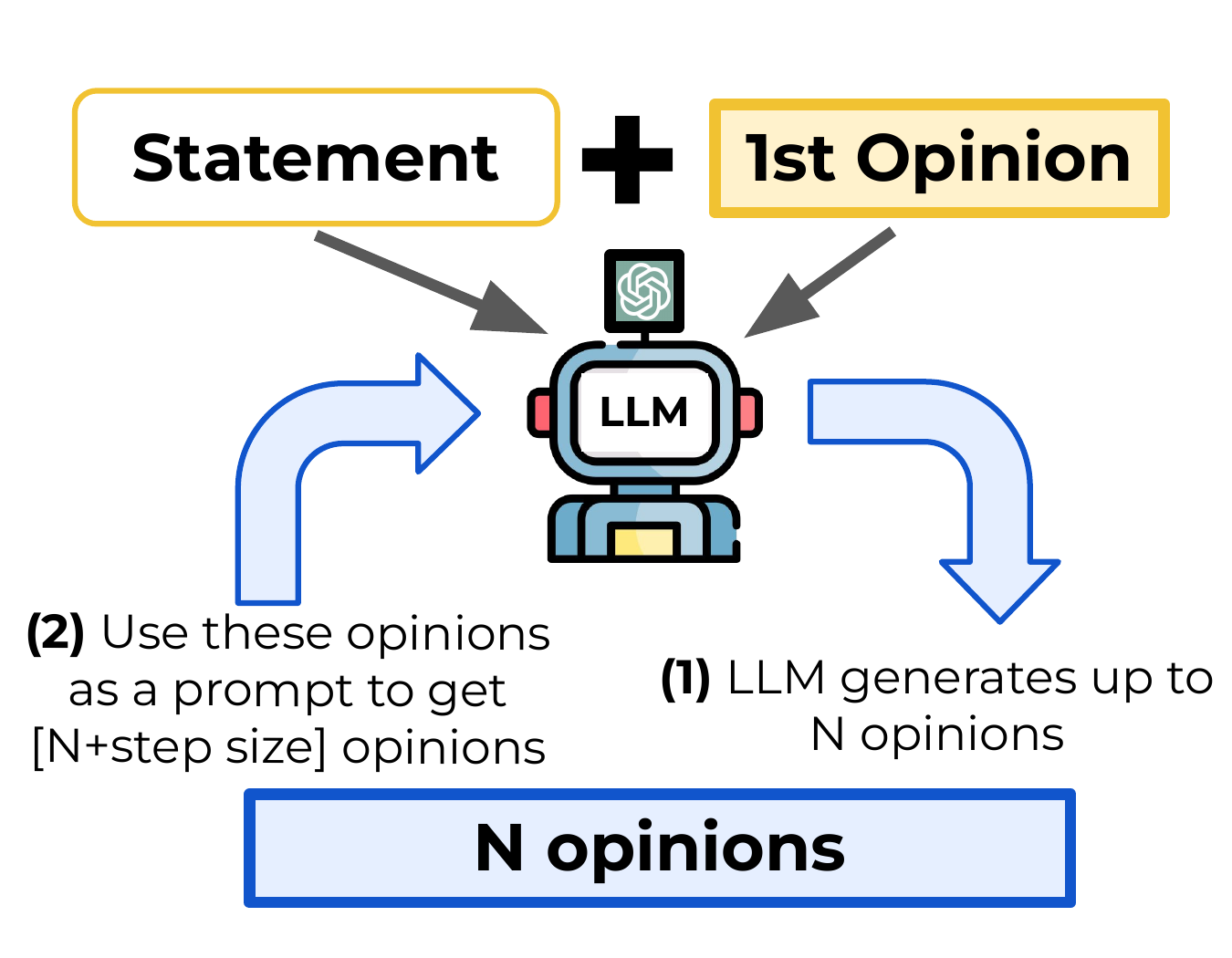}
    \caption{Step-by-step recall prompting. The statement and first generated opinion become the demonstration for prompting the LLM to generate $N$ opinions. The blue-colored parts (Steps 1 and 2) are done incrementally with step size = 3.}
    \label{fig:recall_prompting}
    \vspace{-2mm}
\end{figure}

Once we identify the best setups, we expand our diversity prompting approach to include step-by-step recall prompting to assess the LLMs' diversity coverage. In this experiment, no examples are provided in the prompt. Instead, we only extract one opinion for a given statement and prompt the model to generate additional opinions iteratively until reaching a specified number, $N$ (Figure \ref{fig:recall_prompting}). Without 1-shot demonstration, weaker LLMs often struggle to produce structured outputs. The purpose of this recall prompting experiment is to identify the ``saturation point'' of the model's diversity, which is the maximum number of unique diverse opinions an LLM can generate. 

Across the experiment, we set $N\in\{2, 5, 8, 11, 14, 17, 20\}$.
The first opinion generated by the LLM guides the structured output format since few-shot prompting is not employed in this experiment.

\section{Experimental Setups}

\subsection{Models and Prompting}

We experiment primarily with four GPT variants: GPT-4o\footnote{\url{ https://openai.com/index/hello-gpt-4o/}}, GPT-4, GPT-3.5 \cite{openai2023gpt4}, and GPT-3 (\texttt{text-davinci-002}) \cite{brown2020language}, along with Llama3-70b-chat\footnote{\url{https://llama.meta.com/llama3/}} and Mixtral 8x7B \cite{jiang2024mixtral}. Our baseline involves free-form prompting, where the model generates its stance and reasoning without generating criteria words.
Our primary prompting setup uses in-context learning with one example of ten opinions per statement. We also compare with five-shot prompting across all LLMs and zero-shot prompting only for the strongest model, GPT-4, as weaker models may struggle to generate structured outputs without examples. In step-by-step recall prompting experiments, we vary the number of opinions generated. 

\subsection{Datasets}



\paragraph {\textsc{Social-Chem-101}} contains texts about of social norms and moral judgments for a given situation written by crowd-workers \cite{forbes-etal-2020-social}. 
Since social norms depend on many factors such as the group's beliefs and cultures \cite{ajzen1991theory, 3d9dc9d9-9e8f-3400-ae26-6e37d3097c38}, this dataset is suitable for our task of maximizing LLM's diversity capability. For our study, we randomly sampled 500 texts for the criteria-based vs. free-form prompting experiment and 200 texts for the step-by-step recall prompting.

\paragraph{\textsc{Change My View (CMV)}} consists of debates from online forum threads of the subreddit \textit{Change My View} collected by \citet{hidey-etal-2017-analyzing}. We only take the title of each discussion since it is usually the claim of the argument, resulting in a total of 67 unique claims. 
We use this dataset to examine if LLMs can produce diverse opinions on a highly subjective task because an argumentation task could be highly controversial \cite{537a0c395cd84a88aae06da7673872ff}. 

\paragraph{\textsc{Hate Speech}} dataset contains texts categorized as either ``hate'' or ``not hate'' speech.
\cite{vidgen-etal-2021-learning}. From this dataset, we randomly sample 200 instances, focusing only on implicit hate speech texts which are harder to detect. Hate speech detection is a subjective task because annotators' background may affect how they rate the hate speech label of a text \cite{sap-etal-2019-risk, ghosh-etal-2021-detecting}. We add this task for the step-by-step recall prompting experiment to show how criteria-based prompting can be applied to subjective labeling problems. 

\paragraph{\textsc{Moral Stories}} is a crowd-sourced narrative story dataset \cite{emelin-etal-2021-moral}. For this study, the LLM needs to continue the story with the situation part as. We also randomly sample 200 instances from this dataset and use this dataset for the step-by-step recall prompting experiment to show how our prompting method can be applied to open-ended generation problems.

\subsection{Evaluation Metric}
\label{eval_metric}
\paragraph{Semantic diversity} 
To examine the semantic diversity of the model's reasons using both criteria-based and free-form prompting, we convert the LLM-generated ``reasons'' for each statement into sentence embeddings using SentenceBERT \cite{reimers-gurevych-2019-sentence} with DistilRoberta \cite{sanh2019distilbert}. Next, we compute the cosine distance between every pair of reasons and calculate the average cosine distance across all pairs to measure the statement's semantic diversity score. We average these scores across all statements to obtain the overall semantic diversity.

\begin{table*}[h]
\centering
\begin{tabular}{@{}l|l |r r | r r @{}}
\toprule
& & \multicolumn{2}{c|}{\textsc{Social-Chem-101} ($\uparrow$)} & \multicolumn{2}{c}{\textsc{CMV} ($\uparrow$)}\\
\textbf{Model} &  \textbf{\#Parameters} &\textbf{Free-form} &  \textbf{Criteria} &   \textbf{Free-form} &  \textbf{Criteria}\\
\midrule
GPT-4 & - & 0.3883 & \colorbox{pink!30}{\textbf{0.3919}} & 0.3701 & \colorbox{pink!30}{\textbf{0.3776}}\\
GPT-4o & - & 0.3525  & \colorbox{pink!30}{0.3545} & 0.3480 & \colorbox{pink!30}{0.3759*}\\
GPT-3.5 & - & 0.2865 & \colorbox{pink!30}{0.3100*} & 0.2368 & \colorbox{pink!30}{0.2829*}\\
GPT-3 & 175B& 0.1947 & \colorbox{pink!30}{0.2673*} & 0.1533 & \colorbox{pink!30}{0.2046*}\\
Llama3-chat & 70B & 0.3152 & \colorbox{pink!30}{0.3196} & \colorbox{yellow!30}{0.3158} & 0.3115\\
Mixtral & 46.7B & 0.2657 & \colorbox{pink!30}{0.3186*} & 0.1345 & \colorbox{pink!30}{0.1908*}\\ \hline
Zero-shot GPT-4 & - & \colorbox{yellow!30}{0.3176} & 0.2885 & \colorbox{yellow!30}{0.2669} & 0.2410\\
\bottomrule
\end{tabular}
\caption{Semantic diversity (cosine distance) results on criteria-based prompting vs. free-form prompting experiments. Both setups use one-shot learning except for ``zero-shot GPT-4.'' \textbf{One-shot criteria-based prompting} generally generates more diverse opinions (\colorbox{pink!30}{pink box}) with GPT-4 performing the best. *p < 0.01
}
\label{tab:new_diversity_result}
\end{table*}

\begin{figure*}[h!]
    \centering    
    \begin{subfigure}{0.4\textwidth}
        \centering
        \includegraphics[width=0.95\linewidth]{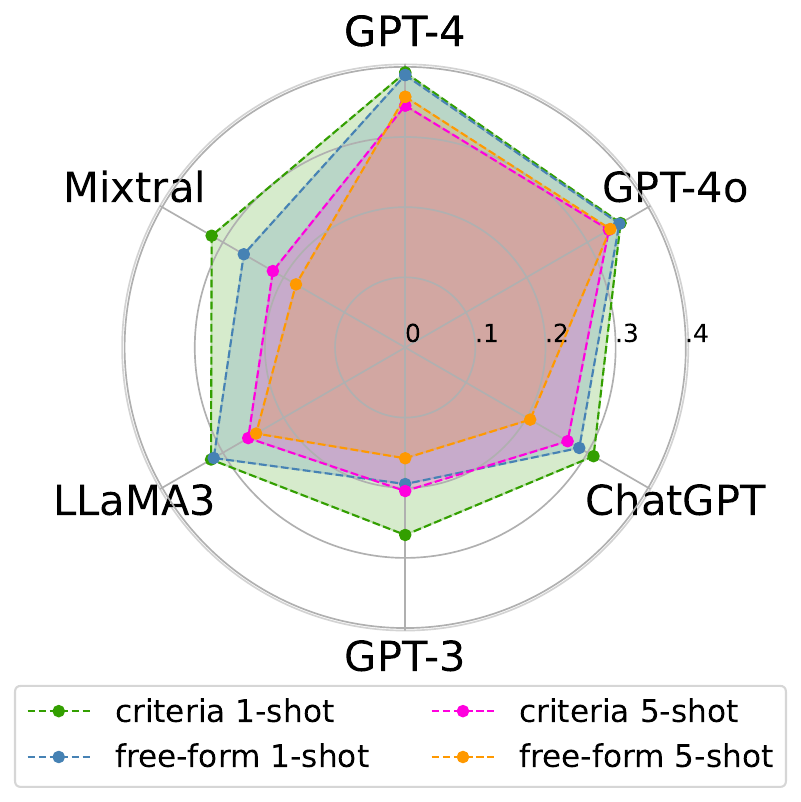}
        \caption{\textsc{Social-Chem-101}}
        \label{fig:sochem_radar}
    \end{subfigure}
    \begin{subfigure}[b]{0.4\textwidth}
        \centering
        \includegraphics[width=0.95\linewidth]{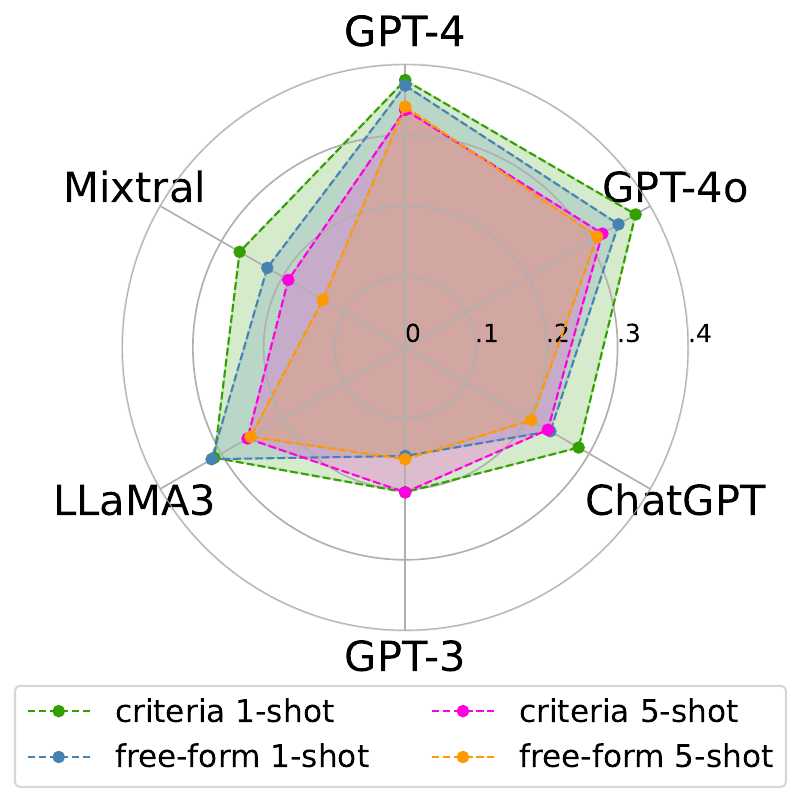}
       \caption{CMV}
       \label{fig:cmv_radar}
    \end{subfigure}
    \caption{
    Semantic diversity score for different LLMs and prompting methods for \textsc{Social-Chem-101} (left) and \textsc{CMV} (right) datasets. \textbf{Criteria-based prompting is the best diversity extraction method for across LLM variants, datasets, and various shots.} We also found that \textbf{too many examples may hurt diversity (5-shot results).}  The results on \textsc{Social-Chem-101} are statistically significant with p < 0.05 (GPT-4) and p < 0.01 (the rest of the models) and p < 0.01 for GPT-3 and Mixtral for CMV.}
    \label{fig:radar}
    \vspace{-5mm}
\end{figure*}

\paragraph{Perspective diversity} \label{eval_metric:perspective}

To evaluate perspective diversity in step-by-step recall prompting, we utilize criteria words generated by LLMs. Some words with similar meanings can be conveyed in different ways. For instance, given a statement ``\textit{It is expected that friends will enjoy being around each other},'' the model could generate two opinions; an opinion may contain ``joy'' as one of the criteria while the other opinion contains ``happiness.'' We prompt GPT-4 with 3 examples to cluster criteria words with similar meanings into groups (details in \ref{clustering_prompt}). Two human annotators manually inspect 1,159 clusters of criteria words from 100 randomly sampled statements across \textsc{Social-Chem-101}, \textsc{CMV}, \textsc{Hate Speech}, and \textsc{Moral Stories} (25 statements per dataset). From this study, the annotators agree that an average of 80.95\% of those clusters of criteria words have similar meanings with inter-annotator percentage agreement of 88.85\%. To measure perspective diversity, we count the number of unique criteria clusters for each opinion on a given statement. A higher count indicates greater diversity in the generated opinions.

\section{Experiment Results with Automatic Evaluation}
\subsection{Semantically Diverse Opinions about Social Norms and Argumentation} 
\label{semantic_diversity_result}

\begin{figure*}[h!]
    \centering
    \begin{subfigure}{0.24\textwidth}
        \centering
        \includegraphics[width=\linewidth]{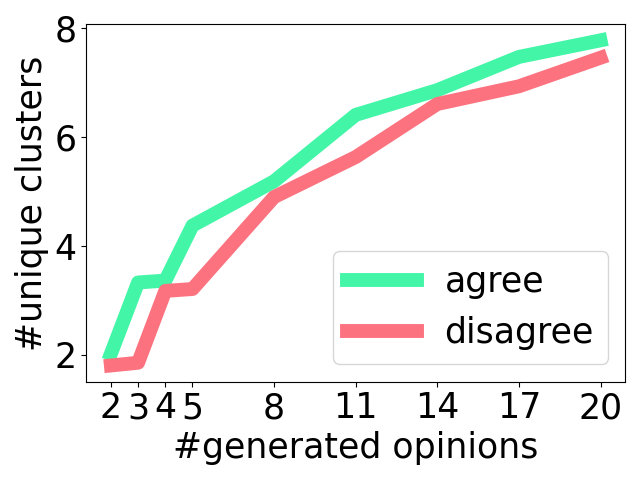}
        \caption{\textsc{Social-Chem-101}}
        \label{fig:sochem_recall_count}
    \end{subfigure}
    \begin{subfigure}[b]{0.24\textwidth}
        \centering
        \includegraphics[width=\linewidth]{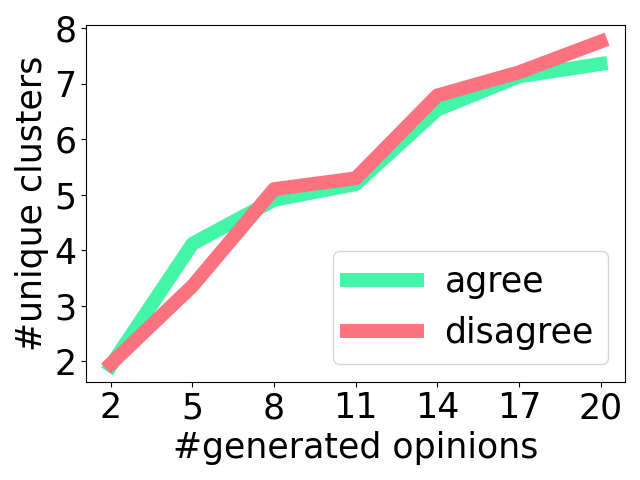}
       \caption{CMV}
        \label{fig:cmv_recall_count}
    \end{subfigure}
    \begin{subfigure}[b]{0.24\textwidth}
        \centering
        \includegraphics[width=\linewidth]{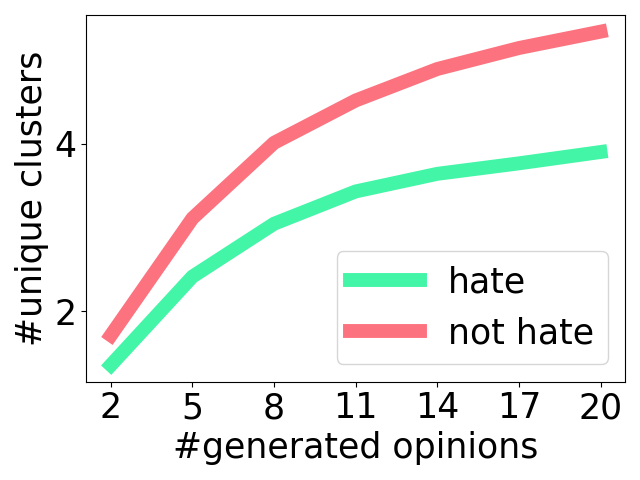}
       \caption{Hate Speech}
        \label{fig:hs_recall_count}
    \end{subfigure}
    \begin{subfigure}[b]{0.24\textwidth}
        \centering
        \includegraphics[width=\linewidth]{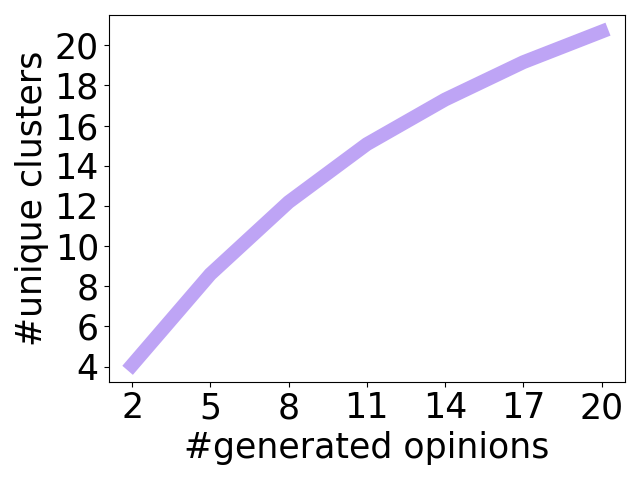}
       \caption{Moral Stories}
        \label{fig:ms_recall_count}
    \end{subfigure}
    \caption{
    X-axis = the number of generated opinions for our diversity coverage experiment. Y-axis = the average number of unique criteria clusters for all statements. Moral Stories do not have stances, so the line is only for all generated continued stories. \textbf{The more subjective a task is,  the more LLM can generate unique criteria clusters.}
    }
    \label{fig:recall_count}
    \vspace{-4mm}
\end{figure*}

Table \ref{tab:new_diversity_result} presents our experiment findings regarding semantic diversity. \textbf{One-shot prompting on GPT-4 produces notably more semantically diverse reasons compared to other models.} Interestingly, weaker models like Mixtral and GPT-3 benefit most from having criteria to guide them toward generating semantically diverse opinions. 

When we prompt GPT-4 without examples and ask for structured outputs only, it tends to generate more diverse reasons with free-form prompting. This aligns with our pilot study where asking criteria without examples proved challenging for the model. Notably, GPT-4 generates an average of 6.8 opinions with zero-shot prompting, fewer than all models with one-shot prompting (GPT-4: 9.9 opinions, Llama3: 10.0, Mixtral: 9.0, GPT-4o: 10.1).

Figure \ref{fig:radar} shows that \textbf{criteria-based prompting consistently outperforms free-form prompting across datasets.} One-shot prompting particularly outputs higher diversity scores than five-shot prompting. This indicates that in five-shot prompting, models tend to adhere more closely to given examples, resulting in less diverse outputs compared to the one-shot setting. Interestingly, in the zero-shot setting, free-form prompting on GPT-4 generates more semantically diverse perspectives than criteria-based prompting. We hypothesize that this occurs because introducing ``criteria'' in the prompt without concrete examples may confuse the model. As a result, the model produces fewer diverse diverse perspectives compared to the simpler free-form prompting.

\subsection{Diversity Coverage by Step-by-Step Recall Prompting} 
Figure \ref{fig:recall_count} shows an increase in the number of unique criteria clusters as the step size increases for the recall step-by-step experiment. For \textsc{Social-Chem-101} and \textsc{CMV}, the model on average generates 8 unique criteria clusters for agreeing and 7 for disagreeing opinions. For \textsc{Hate Speech}, the average number of unique criteria clusters is lower (4 for `hate' and 5 for `not hate'). This demonstrates that labeling hate speech is less subjective compared to social norms (\textsc{Social-Chem-101}) or argumentation (\textsc{CMV}). On the other hand, \textsc{Moral Stories} shows a different trend, with the model generating an average of 20 unique criteria clusters. In Table \ref{table:example:cmv-sc}, we can see examples of opinions generated per statement for the recall prompting experiment. 

\begin{table}[]
    \centering
    \small
    \begin{tabular}{l|l l l}
    \toprule
     \textbf{Task Type} &  \textbf{Dataset} & \textbf{Max} & \textbf{Median}\\ \midrule
     Stance & \multicolumn{3}{l}{\textsc{Social-Chem-101}}\\
     &  Agree  & 17 & 8\\
    & Disagree & 16 & 7\\ \midrule
    Stance & \multicolumn{3}{l}{\textsc{CMV}}\\
     & Agree & 17 & 7\\
    & Disgree & 14 & 7\\ \midrule
    Labeling & \multicolumn{3}{l}{\textsc{Hate Speech}}\\
   & Hate & 14 & 4\\
     & Not Hate & 16 & 5\\ \midrule
    Generation & \multicolumn{3}{l}{\textsc{Moral Stories}}\\
     & All & 47 & 20\\
    \bottomrule
    
    \end{tabular}
    \caption{Different task types with the maximum and the median of the number of unique criteria clusters. More details are in Figure \ref{fig:detailed_recall_hist}.}
    \label{tab:summary_of_diversity_coverage}
    \vspace{-5mm}
\end{table}

Since we limit the maximum number of generated opinions ($N = 20$), the lines on the graph in Figure \ref{fig:recall_count} may seem not to have converged yet. 
However, increasing $N$ does not necessarily lead to a higher number of unique clusters as this is task-dependent (see Table \ref{tab:summary_of_diversity_coverage}). For instance, in the hate speech labeling task, although we set a maximum $N=20$, the highest number of unique clusters is 14 for both ``hate'' and ``not hate," with average number of unique clusters is 4 and 5, respectively. For open-ended problems such as story continuation (\textsc{Moral Stories}), increasing $N$ would lead to a greater number of unique clusters.

\begin{table*}[ht!]
\resizebox{\textwidth}{!}{
\begin{tabular}{@{}ll|ll@{}}
\toprule
\multicolumn{2}{c|}{\textbf{Statement}} &
  It's good to be a hard worker. &
   \\ \cline{1-3}
   \multicolumn{1}{l|}{\multirow{12}{*}{\textbf{GPT-4}}} &
  \multicolumn{1}{l|}{\multirow{6}{*}{\textit{Agree}}} &
  \begin{tabular}[c]{@{}l@{}}1. Being a hard worker increases productivity and pushes one towards success and achieving their goals. \end{tabular} \\ 
  \multicolumn{1}{l|}{} &
  \multicolumn{1}{l|}{} &
  \quad\textcolor{cyan}{Criteria}: \textbf{productivity}, \textbf{success}, \textbf{goals}\\ 
  \multicolumn{1}{l|}{} &
  \multicolumn{1}{l|}{} &\begin{tabular}[c]{@{}l@{}}2. Being a hard worker can bring a sense of fulfillment, achievement, and boost self-esteem. \end{tabular} \\ 
  \multicolumn{1}{l|}{} &
  \multicolumn{1}{l|}{} &
  \quad\textcolor{cyan}{Criteria}: \textbf{fulfillment, achievement,} \textbf{self-esteem}\\ 
  \multicolumn{1}{l|}{} &
  \multicolumn{1}{l|}{} &\begin{tabular}[c]{@{}l@{}} 3. Being a hard worker develops resilience, stokes determination and builds hardiness. \end{tabular}\\ 
  \multicolumn{1}{l|}{} &
  \multicolumn{1}{l|}{} &
  \quad\textcolor{cyan}{Criteria}: \textbf{resilience}, \textbf{determination}, \textbf{ hardiness}\\ \cline{2-3}
  \multicolumn{1}{l|}{} &
  \multicolumn{1}{l|}{\multirow{6}{*}{\textit{Disagree}}} &
  \begin{tabular}[c]{@{}l@{}} 1. Being a hard worker can damage work-life balance, cause burnout and high levels of stress.\end{tabular} \\ 
\multicolumn{1}{l|}{} & \multicolumn{1}{l|}{} & 
\quad\textcolor{cyan}{Criteria}: \textbf{work-life balance},  \textbf{burnout}, \textbf{stress}                                                                                 \\  
\multicolumn{1}{l|}{} & \multicolumn{1}{l|}{} & \begin{tabular}[c]{@{}l@{}}
  2. Being a hard worker can decrease leisure time, family time and negatively affect health. \end{tabular}    \\ 
\multicolumn{1}{l|}{} & \multicolumn{1}{l|}{} & 
\quad\textcolor{cyan}{Criteria}: \textbf{leisure}, \textbf{family}, \textbf{health}                                                                         \\ 
\multicolumn{1}{l|}{} & \multicolumn{1}{l|}{} &  \begin{tabular}[c]{@{}l@{}} 3. Being a hard worker can prevent peace of mind, disrupt contentment, and break restfulness. \end{tabular}                 \\ 
\multicolumn{1}{l|}{} & \multicolumn{1}{l|}{} & 
\quad\textcolor{cyan}{Criteria}: \textbf{peace of mind}, \textbf{contentment},    \textbf{restfulness}       \\ \midrule\midrule
\multicolumn{1}{l|}{\multirow{6}{*}{\textbf{Human}}} &
  \multicolumn{1}{l|}{\multirow{3}{*}{\textit{Agree}}} &
  \begin{tabular}[c]{@{}l@{}} 1. Being hard worker cultivates discipline.\end{tabular}\\
   
  \multicolumn{1}{l|}{} &
  \multicolumn{1}{l|}{} &\begin{tabular}[c]{@{}l@{}}
  2. Hard work often leads to personal growth and development.\end{tabular} \\ 
  \multicolumn{1}{l|}{} &
  \multicolumn{1}{l|}{} &\begin{tabular}[c]{@{}l@{}}
    3. Hard workers tend to be more reliable. \end{tabular}\\ 
 \cline{2-3}
  \multicolumn{1}{l|}{} &
  \multicolumn{1}{l|}{\multirow{3}{*}{\textit{Disagree}}} &
  \begin{tabular}[c]{@{}l@{}}1. Being overly focused on hard work can lead to burnout. \end{tabular} \\ 

\multicolumn{1}{l|}{} & \multicolumn{1}{l|}{} & \begin{tabular}[c]{@{}l@{}} 2. Sometimes working smart is more effective than working hard. \end{tabular}  \\ 
 
\multicolumn{1}{l|}{} & \multicolumn{1}{l|}{} &  \begin{tabular}[c]{@{}l@{}} 3. The value of hard work can very depending on the context.  \end{tabular}                 \\  
\bottomrule
\end{tabular}%
}

\caption{Opinions generated by GPT-4 (top) and a human (bottom) about a statement from \textsc{Social-Chem-101}.}
\label{table:example:cmv-sc}
\end{table*}

\begin{table}
\resizebox{0.48\textwidth}{!}{%
\begin{tabular}{cccc}

\toprule
  &                & \textsc{Social-Chem-101} & \textsc{CMV}  \\ \hline
\multirow{3}{*}{\textit{Agree}}    & \textbf{Human} &  9.17 \small{$\pm 3.16$}    &  10.56\small{$\pm 3.86$}  \\
\cline{2-4} 
   & \textbf{GPT-4} &  8.14\small{$\pm 2.40$}            & 7.86 \small{$\pm 2.62$} \\ \midrule\midrule
   
\multirow{3}{*}{\textit{Disagree}} & \textbf{Human} &  10.04 \small{$\pm 3.31$}            & 11.00\small{$\pm 3.81$} \\ \cline{2-4} 
 & \textbf{GPT-4} & 7.91  \small{$\pm 2.60$}          & 8.30 \small{$\pm 2.74$}\\ \bottomrule
\end{tabular}%
}
\caption{Average number of criteria clusters of human opinions vs. GPT-4-generated opinions per statement with standard deviation. \textbf{While humans can write more diverse opinions when asked, LLM's capability for extracting diverse perspectives is quite on par with human capability.}
}
\label{table:task4-stat}
\vspace{-4mm}
\end{table}

\section{Human vs. LLMs: Diverse Opinion Generation} \label{sec:human-vs-llm}

To assess human capabilities of generating diverse opinions, we hire crowd-workers from Amazon Mechanical Turk to generate as many opinions as they can ($\ge 3$) for each stance (agree, disagree) on 100 statements from \textsc{Social-Chem-101} and 67 statements from \textsc{CMV}. These statements are a subset of the dataset used in our recall prompting experiments. Each worker writes opinions for five statements per HIT, compensated at \$2 per HIT. 

For each human-written opinion, we query GPT-4 to extract criteria words and cluster them using the same method employed for computing perspective diversity of model-generated opinions. Table \ref{table:task4-stat} shows that \textbf{humans tend to produce slightly more diverse opinions than LLMs}, with approximately 1 or 2 more criteria for \textsc{Social-Chem-101} and 3 more criteria for \textsc{CMV}. 

\begin{figure}[t!]
    \centering
    \hspace*{-0.5cm}
    \includegraphics[width=1.1\columnwidth,trim={0.2cm 0.1cm 0.1cm 0.1cm},clip]{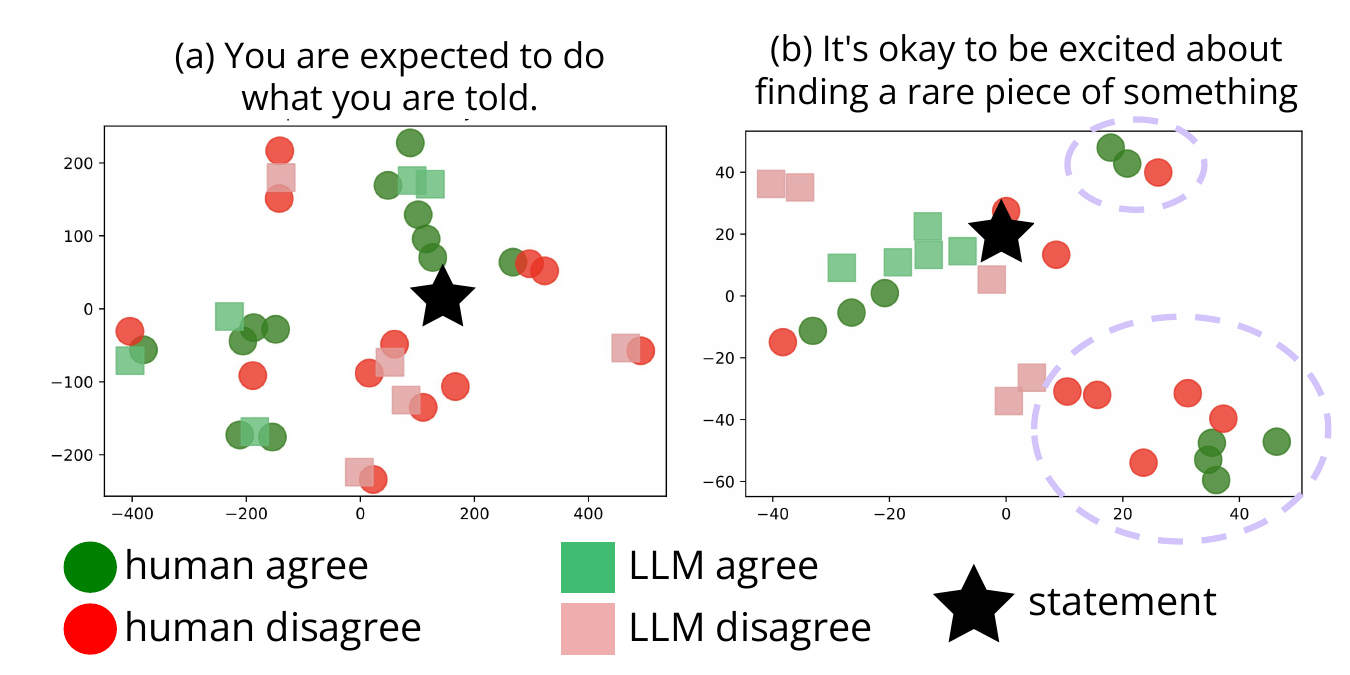}
    \caption{T-SNE for opinions written by human and generated by GPT-4. \textbf{LLM can mostly generate both agree and disagree opinions that align with human when they are semantically close to the statement}.}
    \label{fig:tsne_human_model}
    \vspace{-5mm}
\end{figure}

Figure \ref{fig:tsne_human_model} illustrates two statements alongside their respective opinions by humans and GPT-4 in T-SNE plot. The statements and opinions are embedded with the same approach for semantic diversity experiment. We observe that \textbf{LLMs can generate agreeing and disagreeing opinions that align with human perspectives}, despite LLMs producing slightly fewer opinions. The failure cases of LLMs occur when human opinions diverge semantically from the statement (e.g., the lower right under purple circles).

\begin{figure*}[]
    \centering
\includegraphics[width=0.9\linewidth]{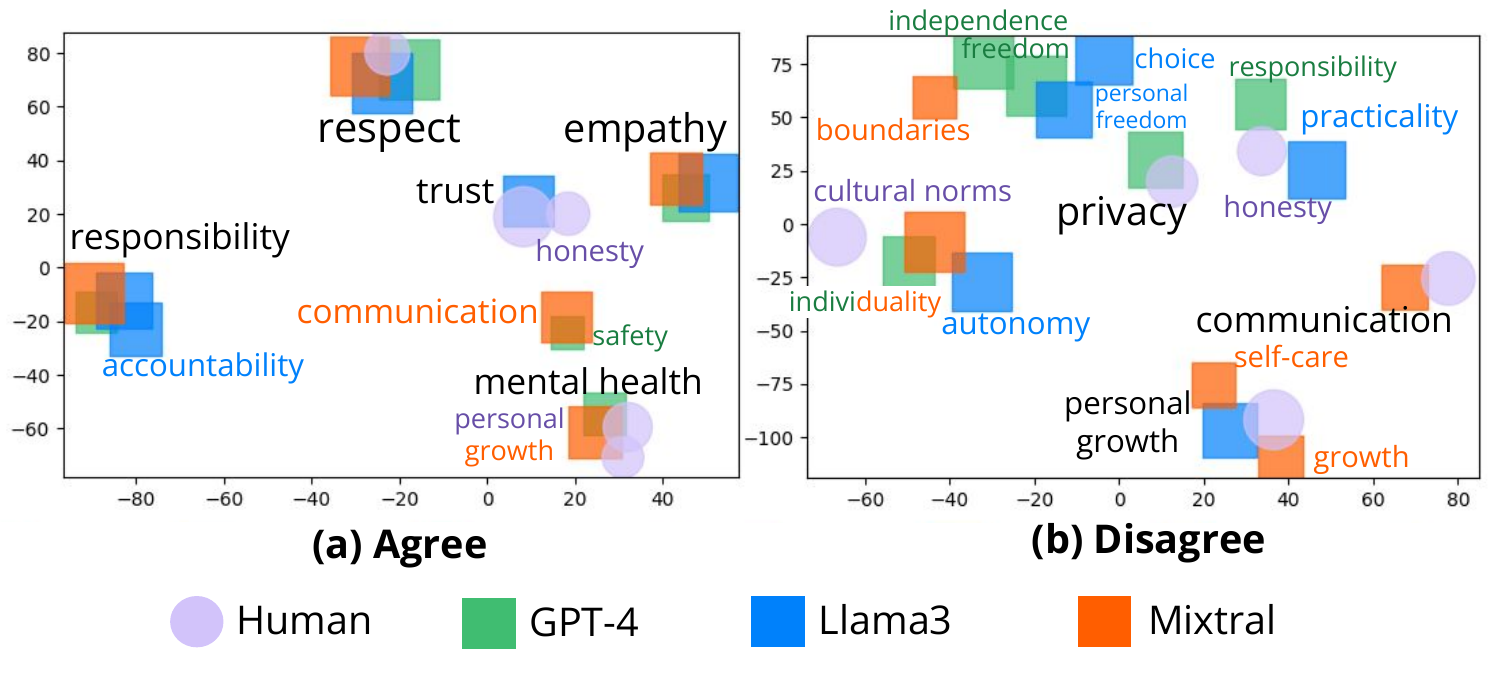}
    \caption{T-SNE plots of five most frequent criteria words by humans and three LLMs: GPT-4, Llama3, and Mixtral. The size of a point represents the frequency. Black font refers to the label of all points next to the text. \textcolor{thepurple}{Purple} font for human's criteria words, \textcolor{darkgreen}{green} for GPT-4, \textcolor{myblue}{blue} for Llama3, and \textcolor{orange}{orange} for Mistral. \textbf{LLMs generally mimic human values, although at times they tend to regard rule-following notions as important (e.g., ``responsibility'' or ``safety'') for agree opinions and extreme freedom (e.g., ``independence,'' ``boundaries,'' ``individuality'') for disagreeing opinions more than humans do.}
    \vspace{-4mm}
    }
    \label{fig:criteria_words_tsne}
\end{figure*}

\paragraph{Criteria Words by Different LLMs and Humans} 
We analyze the frequent criteria words generated by GPT-4, Llama3, Mixtral, and humans using T-SNE embeddings in Figure \ref{fig:criteria_words_tsne} for agreeing and disagreeing opinions in \textsc{Social-Chem-101}. From each model and humans, we select the top 5 frequent criteria words across all statements and visualize their embeddings on a T-SNE plot.

For agreeing opinions, in general the three LLMs quite align with humans. GPT-4 and Llama3 have ``respect'' as the most frequent criterion, and all three LLMs regard ``responsibility,''  ``safety,'' and ``emapthy'' as important criteria. Meanwhile, humans value ``trust'' the most, and only Llama3 aligns with human for this value. For the disagreeing opinions, we can see that humans value ``personal growth'' the most and then followed by ``cultural norms,'' ``communication,'' ``privacy,'' and ``honesty.'' However, only Llama3 also considers ``personal growth'' as important. In general, all LLMs consider that ``freedom'' and ``autonomy'' are the most important which sounds more extreme compared to human values.

We also examine how much the criteria words generated by LLMs agree with human responses using top-p sampling (p=10\%). For both agreeing and disagreeing opinions, we found that GPT-4 agrees the most with humans (agree: 45.63\%, disagree: 39.53\%), followed by Llama3 (30.00\% and 28.14\%), and Mixtral (29.38\% and 26.35\%). Since the criteria words by humans are extracted by GPT-4, there may be a lexical bias toward words that GPT-4 frequently uses. However, the T-SNE plot displays the semantic closeness of the criteria words by these different LLMs and humans. All three LLMs align well with humans with a tendency of favoring rule-following criteria words for agreeing opinions and extreme independence for disagreeing opinions. 


\begin{figure}[ht!]
    \centering    
    \begin{subfigure}{0.4\textwidth}
        \centering
        \includegraphics[width=\linewidth]{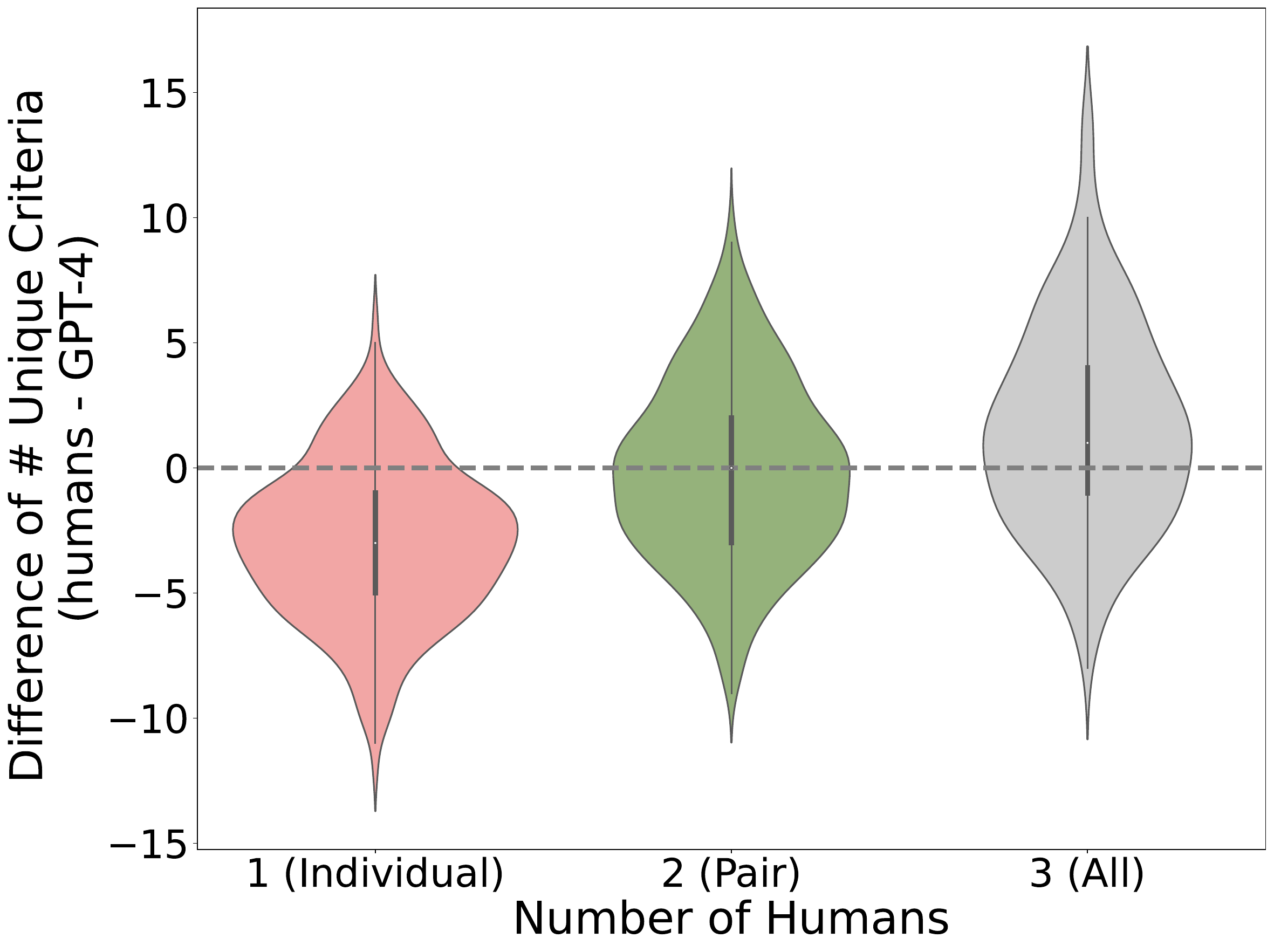}
        \caption{\textsc{Social-Chem-101}}
        \label{fig:sochem_violin}
    \end{subfigure}
    \begin{subfigure}[b]{0.4\textwidth}
        \centering
        \includegraphics[width=\linewidth]{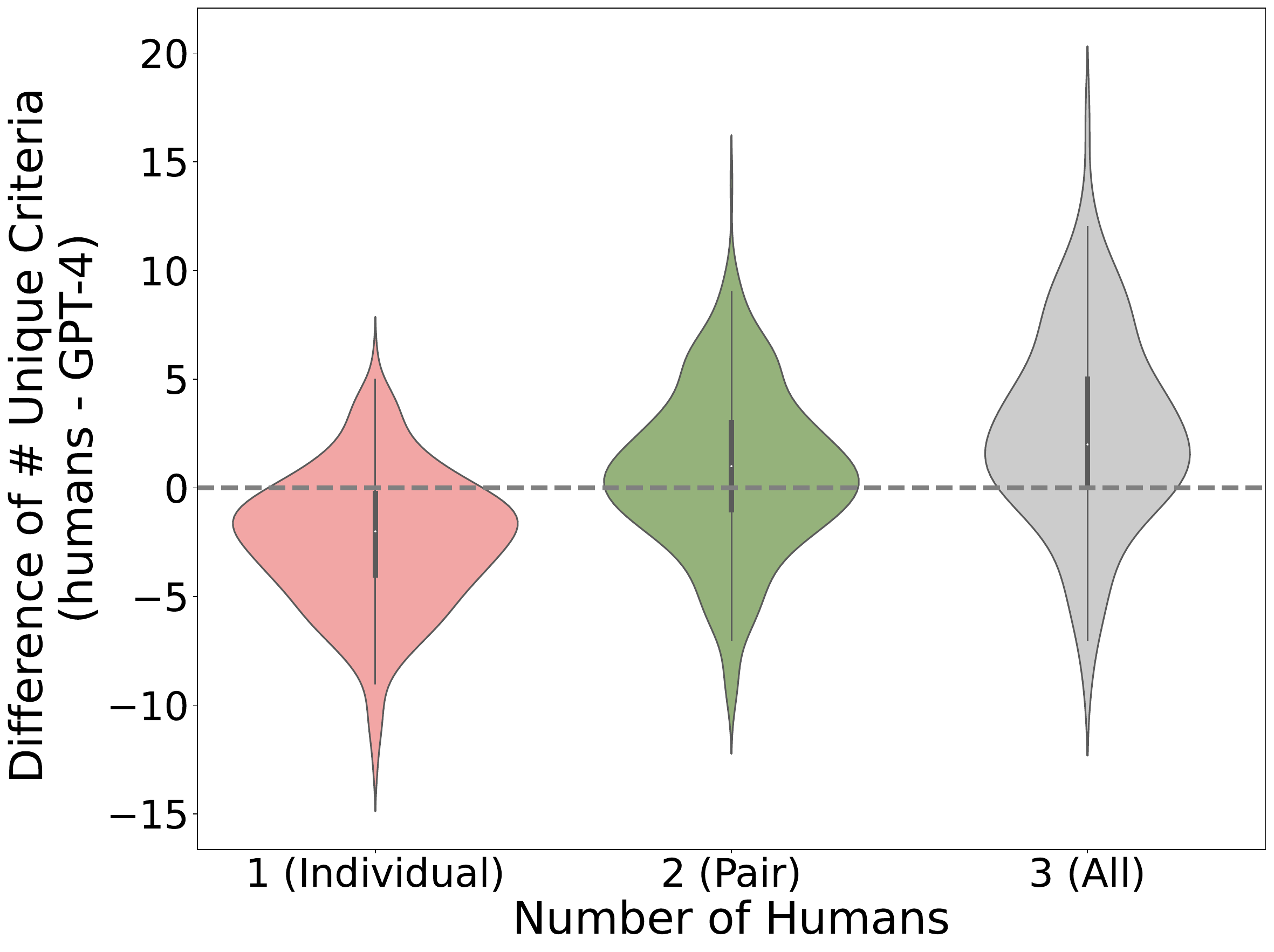}
       \caption{CMV}
       \label{fig:cmv_violin}
    \end{subfigure}
    \caption{
    The distribution of the differences in the number of unique criteria clusters between human and GPT-4. \textbf{A pair of humans can equalize LLM's capability of extracting maximum diversity.}
    }
    \label{fig:violin}
    \vspace{-4mm}
\end{figure}

\paragraph{When Will Human Reach LLM's Diversity Generation Capability?} 
To examine how many humans are needed to match the diversity generation capability of LLMs, we compute the difference of the number of unique clusters of criteria in human opinions vs. GPT-4's generation. Figure \ref{fig:violin} visualizes the distribution of these differences between humans and GPT-4. Our analysis indicates that a person tends to generate fewer unique perspectives compared to GPT-4. However, \textbf{a pair of humans or more shows a higher density of matching or exceeding the diversity capability of GPT-4} when generating unique opinions for statements in both \textsc{Social-Chem-101} and \textsc{CMV}. This suggests the importance of ``communication'' between individuals to broaden one's perspectives, underscoring the value of having an LLM capable of generating diverse viewpoints.

\section{Related Work}
\paragraph{Diversity in NLP}
Diversity in NLP has been extensively explored across various dimensions: (1) lexical variability \cite{dusek-kasner-2020-evaluating, tevet-berant-2021-evaluating, li-etal-2016-diversity}; (2) syntactical diversity \cite{giulianelli-etal-2023-comes, huang-etal-2023-paraamr}; (3) semantic diversity \cite{stasaski-hearst-2022-semantic, reimers-gurevych-2019-sentence, zhang2020bertscore}; and (4) perspective diversity \cite{plank-2022-problem, hayati-etal-2021-bert, santy-etal-2023-nlpositionality}. Some studies focus on annotator diversity \cite{rottger-etal-2022-two, wich-etal-2021-investigating}, while others examine diversity in generated language \cite{hashimoto-etal-2019-unifying, liu2023geval}. Our work aligns closely with prior studies on perspective diversity, specifically in examining stances and rationales generated by LLMs. However, unlike previous research primarily focusing on classification tasks, our investigation encompasses sentence-level reasoning diversity, offering a more nuanced perspective. \citet{joshi-etal-2020-state} argue that NLP research is biased toward Western perspectives. Thus, diverse perspectives from minority populations are relatively overlooked. Our work is important to uncover the extent to which perspective diversity can be extracted from LLMs.

\paragraph{Diversity Generation by LLMs}

LLMs have been utilized extensively to produce diverse synthetic datasets, such as paraphrasing \cite{cegin2023chatgpt}, structured wiki-like bios for notable figures \cite{yuan2022synthbio}, and instruction datasets \cite{wang-etal-2023-self-instruct, alpaca, honovich-etal-2023-unnatural}.

Unlike diverse large-scale data generation, \citet{lahoti2023improving, giulianelli-etal-2023-comes} specifically examine variability in model responses. They propose novel prompting techniques to enhance diversity in LLM outputs, particularly concerning gender and cultural prompts. Our work aligns closely with \citet{lahoti2023improving} by advocating for fairness in LLM outputs through perspective diversity, which goes beyond semantic variability.

Additionally, our approach is similar to \citet{giulianelli-etal-2023-comes} in generating multiple responses per prompt. To promote opinion diversity in LLMs, \citet{aroyo2023dices} introduce a dataset labeled by human raters from different demographic populations, focusing on safety such as bias, misinformation, and harmful content. Despite the previous claim that LLMs can produce diverse content, some studies suggest that co-writing with LLMs may affect human writers' opinions \cite{Jakesch_2023} and reduce writing diversity \cite{padmakumar2024does}. Our research addresses this gap by proposing a method to generate diverse perspectives rather than a single dominant opinion.

\section{Conclusion and Future Work}

To the best of our knowledge, this is the first work that tackles extracting maximum perspective diversity from LLMs. To do this, we propose a criteria-based prompting method and probe LLMs' capacity to generate as many diverse perspectives as possible and explain their reasons for choosing their corresponding stances on subjective statements. Through our step-by-step recall prompting, we characterize the subjectivity of various tasks and reach the maximum diversity of LLM's generation. LLMs can generate comparable number of diverse outputs with humans and similar values as humans' responses. As we compare LLMs' opinion generations with human's, they are quite ``precise'' as they are semantically similar to human opinions, bur their recall is slightly lower than humans.  

While the number of criteria clusters does not precisely mean the ideal maximum diversity, it indicates that we could use LLMs to push further perspective diversity to include more diverse opinions. Our work opens up a wider range of possibilities for examining more advanced diversity ``quantification'' and ``maximization'' methods. There are also many application possibilities for extending this work in the future. In this study, we have not assessed how much the extracted diverse opinions are similar with the real world's diverse opinions yet. Instead, we focus on the diversity coverage. Future work could evaluate this by comparing the distribution of extracted opinions with a distribution of people’s opinions collected from real world survey or poll data. We recommend further exploration of cultural aspects, persona, or human values on diversity extraction. Moreover, our method could be applied for curating diverse data for open-ended tasks, such as generating diverse outputs for instruction-tuning tasks or subjective task labeling. Findings that it takes two humans to equalize LLMs' diverse generation capacity suggest that communications between multiple humans or LLMs can be a future work to introduce more diverse perspectives. 

\section*{Limitations and Ethical Considerations}
While our prompting approach does not generate an exhaustive, complete list of diverse opinions, our study serves as a comparative study that examines the capability of various language models for generating diverse opinions given various numbers of examples and input datasets. Moreover, for now, we only experimented with our proposed criteria-based prompting technique for subjective tasks. It would be interesting future work to try the same technique on non-subjective tasks. Currently, we rely on LLMs (GPT-4) to generate criteria words from non-criteria prompting outputs. Future works could deal with in-depth variations of these criteria word extraction methods and analyses on the words themselves. 

We noticed that the demographics of crowd-workers who participate in the opinion writing are skewed toward white with bachelor degree as their highest education level. Demographic factors, including culture, may impact how these opinions are written. For future work, it would be interesting to compare more opinions written by participants from other cultures with the model's generated opinions. 

We also have not explored all different combinations of setups of decoding parameters besides comparing different temperatures and top$\_$p sampling during the initial experiments. However, we would like to highlight that our work is not simply probing LLM's ability to generate diverse tokens that may convey similar meaning, but rather if the LLM has the capability for generating diverse perspectives. Examining various decoding methods could be a potential future work for this study. 

Potential risks could be a situation where our criteria-based prompting attempts to generate diverse opinions on certain topics that are socially unacceptable and/or contain harmful content. To mitigate any concern regarding this issue, we suggest that researchers carefully review the content of subjects before applying our prompting approach to their work. 

For the human study, our institution determined our study as exempt from IRB review. Since the topic of some statements could trigger human workers, we added warnings before the worker could proceed to work on our task. 

\section*{Acknowledgment}
We would like to thank members of the Minnesota NLP lab and Google reviewers for their valuable feedback and suggestions on the paper draft. 

\bibliography{custom, anthology}

\appendix

\section{Appendix}\label{sec:appendix}
\subsection{Other prompting techniques we tried}
\label{other_prompting}
During our pilot study with a smaller number of subjective statements, we tried various prompting methods, such as (1) zero-shot vs. few-shot prompting; (2) the number of opinions (10, 15, 20) in the few-shot examples; (3) prompts to generate structured output vs. unstructured output; and (4) varying N greater than 20 (e.g., 30, 50, 100) for our step-by-step recall prompting. However, it would be too costly in terms of budget and time to run all the combinations of set-ups, so we chose the current setup since it produces enough diverse output to examine the saturation point and the structured outputs are easy to process. We are aware of other prompting techniques such as Chain-of-Thought (CoT) prompting \cite{wei2023chainofthought}, least-to-most prompting \cite{zhou2023leasttomost}, and self-consistency prompting \cite{wang2023selfconsistency}. However, none of them handle the task of asking LLM to generate the maximum diversity. On the other hand, our diversity prompting can complement these other prompting approaches for extracting diversity. 

\subsection{Prompting}
\label{prompt_details}
\begin{itemize}
    \item Free-form prompting 

    Five-shot free-form prompt is shown as follows. One-shot free-form prompting only takes the first statement (``It's rude to use profanity'') and its related opinions as a demonstration in the prompt. 
    
    \texttt{Statement: It's rude to use profanity.}
    
    \texttt{Tell me opinions about the statement as many as possible from different people with "Agree" or "Disagree" and explain how they have different opinions.}
    
    \texttt{Output:}
    \texttt{\{1: \{"Stance" : "Agree", "Reason": "I was brought up in a family and culture where using profanity was considered impolite and disrespectful, and I have continued to hold that belief in my adult life.", 2: \{"Stance": "Disagree", "Reason": "Profanity can be a form of self-expression, and there are situations where it may be appropriate or even necessary to use it to convey strong emotions or opinions."\}, 3: \{"Stance": "Agree", "Reason": "Using profanity goes against accepted norms of etiquette and social behavior, and it is considered impolite and rude in most social settings."\}, 4: \{"Stance": "Disagree", "Reason": "The appropriateness of profanity depends on the context and the individual's perception, as what may be acceptable in one situation may not be in another."\}, 5: \{"Stance": "Agree", "Reason":"Using profanity can disrupt social harmony and respectful communication, and it is important to maintain a certain level of decorum in interactions."\}, 6: \{"Stance": "Disagree", "Reason": "Individuals should have the freedom to choose their language, and using profanity can be a way to express themselves authentically."\}, 7: \{"Stance": "Disagree", "Reason": "In certain cultures or societies, the use of profanity may be more accepted or even a normal part of everyday language."\}, 8: \{"Stance": "Agree", "Reason": "In a professional environment, the use of profanity can be seen as unprofessional and can negatively impact one's reputation and relationships with colleagues."\}, 9:\{"Stance": "Disagree","Reason": "Profanity can be used in art and creative expression to evoke emotions, challenge societal norms, or convey a specific message."\}, 10:\{"Stance": "Agree", "Reason": "I was taught from a young age that using profanity is rude and disrespectful, and I continue to adhere to those values."\}\}}
    
    \texttt{Statement: It's okay to have privacy.}
    \texttt{Tell me opinions about the statement as many as possible from different people with "Agree" or "Disagree" and explain how they have different opinions.}
    
    \texttt{Output:}
    \texttt{\{1: \{"Stance": "Agree", "Reason": "Having privacy allows individuals to establish personal boundaries and maintain their autonomy."\}, 2: \{"Stance": "Disagree", "Reason": "Lack of privacy can promote transparency and build trust in relationships."\}, 3: \{"Stance": "Agree", "Reason": "Privacy provides a sense of security and protection, especially in personal and sensitive matters."\}, 4: \{"Stance": "Disagree", "Reason": "Reduced privacy enables accountability and facilitates supervision, ensuring responsible behavior."\}, 5: \{"Stance": "Agree", "Reason": "Privacy contributes to personal well-being and mental health by allowing individuals to have space for self-reflection and relaxation."\}, 6: \{"Stance": "Disagree", "Reason": "Limited privacy fosters social connectedness and a sense of community, as individuals share their lives more openly."\}, 7: \{"Stance": "Agree", "Reason": "Privacy fosters intimacy and allows individuals to be vulnerable in personal relationships."\}, 8: \{"Stance": "Disagree", "Reason": "Reduced privacy promotes openness and honesty, as individuals are more transparent about their actions and intentions."\}, 9: \{"Stance": "Agree", "Reason": "Privacy facilitates personal growth and self-discovery, providing a space for individuals to explore their interests and identities."\}, 10: \{"Stance": "Disagree", "Reason": "Limited privacy enhances safety and security by allowing for greater surveillance and monitoring of individuals."\}\}}
    
    \texttt{Statement: It's good to confide in your parents the things that you're struggling with.}
    \texttt{Tell me opinions about the statement as many as possible from different people with "Agree" or "Disagree" and explain how they have different opinions.}
    
    \texttt{Output:}
    
    \texttt{\{1: \{"Stance": "Agree", "Reason": "Confiding in parents can provide emotional support, strengthen the parent-child bond, and allow parents to provide guidance in times of struggle."\}, 2: \{"Stance": "Disagree", "Reason": "Some individuals may fear judgment from their parents or value their privacy, leading them to hesitate in confiding their struggles."\}, 3: \{"Stance": "Agree", "Reason": "Sharing struggles with parents can build trust and create opportunities for collaborative problem-solving and finding solutions."\}, 4: \{"Stance": "Disagree", "Reason": "Some individuals believe in resolving their struggles independently, promoting self-reliance and independence."\}, 5: \{"Stance": "Agree", "Reason": "Confiding in parents allows them to provide empathy, understanding, and emotional support during challenging times."\}, 6: \{"Stance": "Disagree", "Reason": "Communication barriers or a generation gap may hinder effective understanding and make individuals less likely to confide in their parents."\}, 7: \{"Stance": "Agree", "Reason": "Parents, having gone through similar experiences, can offer wisdom and advice based on their own life lessons."\}, 8: \{"Stance": "Disagree", "Reason": "Some individuals may have experienced neglect or unhelpful advice from their parents in the past, making them hesitant to confide in them."\}, 9: \{"Stance": "Agree", "Reason": "Sharing struggles with parents can provide a sense of catharsis and emotional relief by expressing emotions and thoughts."\}, 10: \{"Stance": "Disagree", "Reason": "Some individuals find more comfort in confiding their struggles to friends or other trusted individuals, seeking peer support."\}\}}
    
    \texttt{Statement: You should always try and help people in tough situations.}
    
    \texttt{Tell me opinions about the statement as many as possible from different people with "Agree" or "Disagree" and explain how they have different opinions.}
    
    \texttt{Output:}
    
    \texttt{\{1: \{"Stance": "Agree", "Reason": "Helping others in tough situations is an act of empathy, compassion, and kindness, and it can make a positive impact on their lives."\}, 2: \{"Stance": "Disagree", "Reason": "Helping others in tough situations can put one's personal safety at risk and may violate personal boundaries."\}, 3: \{"Stance": "Agree", "Reason": "As members of society, we have a moral obligation and responsibility to help those in need, especially in tough situations."\}, 4: \{"Stance": "Disagree", "Reason": "In some situations, prioritizing self-preservation and personal priorities may be more important than helping others in tough situations."\}, 5: \{"Stance": "Agree", "Reason": "Helping others in tough situations fosters a sense of community and solidarity, and it can strengthen social bonds."\}, 6: \{"Stance": "Disagree", "Reason": "Helping others in tough situations can enable dependency and may not address the root causes of their problems."\}, 7: \{"Stance": "Agree", "Reason": "Helping others in tough situations is an act of altruism and generosity, and it can bring a sense of fulfillment and purpose."\}, 8: \{"Stance": "Disagree", "Reason": "Helping others in tough situations should respect personal boundaries and require consent, as forcing help on someone can be harmful."\}, 9: \{"Stance": "Agree", "Reason": "Helping others in tough situations is a part of humanitarianism and global responsibility, and it can contribute to a more just and equitable world."\}, 10: \{"Stance": "Disagree", "Reason": "Helping others in tough situations should be done with cultural sensitivity and respect, as imposing one's values and beliefs through helping others can be harmful and disrespectful."\}\}}
    
    \texttt{Statement: You shouldn't let people eat in your car.}
    
    \texttt{Tell me opinions about the statement as many as possible from different people with "Agree" or "Disagree" and explain how they have different opinions.}
    
    \texttt{Output:}
    
    \texttt{\{1: \{"Stance": "Agree", "Reason": "Allowing people to eat in your car can lead to messiness and disrespect for your property."\}, 2: \{"Stance": "Disagree", "Reason": "Allowing people to eat in your car can be a hospitable gesture and make them feel more comfortable."\}, 3: \{"Stance": "Agree", "Reason": "Allowing people to eat in your car can leave unpleasant odors and stains that are difficult to remove."\}, 4: \{"Stance": "Disagree", "Reason": "Allowing people to eat in your car can be convenient and save time, especially when on the go."\}, 5: \{"Stance": "Agree", "Reason": "Allowing people to eat in your car can compromise hygiene and pose health risks, especially if the food is not properly handled or stored."\}, 6: \{"Stance": "Disagree", "Reason": "Allowing people to eat in your car is a personal choice and can depend on the relationship with the person and the circumstances."\}, 7: \{"Stance": "Agree", "Reason": "Allowing people to eat in your car can be a distraction and compromise safety, especially if the driver is also eating."\}, 8: \{"Stance": "Disagree", "Reason": "Allowing people to eat in your car can demonstrate flexibility and adaptability, especially in social situations."\}, 9: \{"Stance": "Agree", "Reason": "Allowing people to eat in your car can decrease its resale value and require more maintenance and cleaning."\}, 10: \{"Stance": "Disagree", "Reason": "Allowing people to eat in your car can be acceptable in certain cultural norms and etiquette, and it depends on the context and situation.\}\}}

    \item Prompt for criteria-based prompting:

    \texttt{Statement: It's rude to use profanity.}
    
    \texttt{Tell me opinions about the statement as many as possible from different people with "Agree" or "Disagree", one-word or one-phrase criteria that is important for their opinions, and explain how they have different opinions.}
    
    \texttt{Output:}
    \texttt{\{1: \{"Stance" : "Agree", "Criteria": ["impolite", "disrespectful", "culture", "family"], "Reason": "I was brought up in a family and culture where using profanity was considered impolite and disrespectful, and I have continued to hold that belief in my adult life.", 2: "Stance": "Disagree", "Criteria": ["self-expression", situation], "Reason": "Profanity can be a form of self-expression, and there are situations where it may be appropriate or even necessary to use it to convey strong emotions or opinions.", 3: \{"Stance": "Agree", "Criteria":["etiquette", "social norms", "impolite", "rude"], "Reason": "Using profanity goes against accepted norms of etiquette and social behavior, and it is considered impolite and rude in most social settings."\}, 4: \{"Stance": "Disagree", "Criteria": ["contextual appropriateness", "individual perception"], "Reason": "The appropriateness of profanity depends on the context and the individual's perception, as what may be acceptable in one situation may not be in another."\}, 5: \{"Stance": "Agree", "Criteria": ["harmony", "responsible communication"], "Reason":"Using profanity can disrupt social harmony and respectful communication, and it is important to maintain a certain level of decorum in interactions."\}, 6:\{"Stance": "Disagree", "Criteria": ["individuality", "choice"], "Reason": "Individuals should have the freedom to choose their language, and using profanity can be a way to express themselves authentically."\}, 7:\{"Stance": "Disagree", "Criteria": ["culture", "society"], "Reason": "In certain cultures or societies, the use of profanity may be more accepted or even a normal part of everyday language."\},8:\{"Stance": "Agree", "Criteria": ["professionalism", "workplace"], "Reason": "In a professional environment, the use of profanity can be seen as unprofessional and can negatively impact one's reputation and relationships with colleagues."\},9:\{"Stance": "Disagree", "Criteria": ["artistic expression", "creativity"], "Reason": "Profanity can be used in art and creative expression to evoke emotions, challenge societal norms, or convey a specific message."\},10:\{"Stance": "Agree", "Criteria": ["education", "upbringing"], "Reason": "I was taught from a young age that using profanity is rude and disrespectful, and I continue to adhere to those values."\}\}}
    
    \texttt{Statement: It's okay to have privacy.}
    
    \texttt{Tell me opinions about the statement as many as possible from different people with "Agree" or "Disagree", one-word or one-phrase criteria that is important for their opinions, and explain how they have different opinions.}
    
    \texttt{Output:}
    
    \texttt{\{1: \{"Stance": "Agree", "Criteria": ["personal boundaries", "autonomy"], "Reason": "Having privacy allows individuals to establish personal boundaries and maintain their autonomy."\}, 2: \{"Stance": "Disagree", "Criteria": ["transparency", "trust"], "Reason": "Lack of privacy can promote transparency and build trust in relationships."\}, 3: \{"Stance": "Agree", "Criteria": ["security", "protection"], "Reason": "Privacy provides a sense of security and protection, especially in personal and sensitive matters."\}, 4: \{"Stance": "Disagree", "Criteria": ["accountability", "supervision"], "Reason": "Reduced privacy enables accountability and facilitates supervision, ensuring responsible behavior."\}, 5: \{"Stance": "Agree", "Criteria": ["mental health"], "Reason": "Privacy contributes to personal well-being and mental health by allowing individuals to have space for self-reflection and relaxation."\}, 6: \{"Stance": "Disagree", "Criteria": ["social connectedness", "community"], "Reason": "Limited privacy fosters social connectedness and a sense of community, as individuals share their lives more openly."\}, 7: \{"Stance": "Agree", "Criteria": ["intimacy", "vulnerability"], "Reason": "Privacy fosters intimacy and allows individuals to be vulnerable in personal relationships."\}, 8: \{"Stance": "Disagree", "Criteria": ["openness", "honesty"], "Reason": "Reduced privacy promotes openness and honesty, as individuals are more transparent about their actions and intentions."\}, 9: \{"Stance": "Agree", "Criteria": ["personal growth", "self-discovery"], "Reason": "Privacy facilitates personal growth and self-discovery, providing a space for individuals to explore their interests and identities."\}, 10: \{"Stance": "Disagree", "Criteria": ["safety", "security"], "Reason": "Limited privacy enhances safety and security by allowing for greater surveillance and monitoring of individuals."\}\}\}}
    
    \texttt{Statement: It's good to confide in your parents the things that you're struggling with.}
    
    \texttt{Tell me opinions about the statement as many as possible from different people with "Agree" or "Disagree", one-word or one-phrase criteria that is important for their opinions, and explain how they have different opinions.}
    
    \texttt{Output:}
    
    \texttt{\{1: \{"Stance": "Agree", "Criteria": ["support", "emotional bond", "guidance"], "Reason": "Confiding in parents can provide emotional support, strengthen the parent-child bond, and allow parents to provide guidance in times of struggle."\}, 2: \{"Stance": "Disagree", "Criteria": ["judgment", "privacy"], "Reason": "Some individuals may fear judgment from their parents or value their privacy, leading them to hesitate in confiding their struggles."\}, 3: \{"Stance": "Agree", "Criteria": ["trust", "problem-solving"], "Reason": "Sharing struggles with parents can build trust and create opportunities for collaborative problem-solving and finding solutions."\}, 4: \{"Stance": "Disagree", "Criteria": ["self-reliance", "independence"], "Reason": "Some individuals believe in resolving their struggles independently, promoting self-reliance and independence."\}, 5: \{"Stance": "Agree", "Criteria": ["empathy", "understanding"], "Reason": "Confiding in parents allows them to provide empathy, understanding, and emotional support during challenging times."\}, 6: \{"Stance": "Disagree", "Criteria": ["communication barriers", "generation gap"], "Reason": "Communication barriers or a generation gap may hinder effective understanding and make individuals less likely to confide in their parents."\}, 7: \{"Stance": "Agree", "Criteria": ["shared experiences", "wisdom"], "Reason": "Parents, having gone through similar experiences, can offer wisdom and advice based on their own life lessons."\}, 8: \{"Stance": "Disagree", "Criteria": ["neglect", "unhelpful advice"], "Reason": "Some individuals may have experienced neglect or unhelpful advice from their parents in the past, making them hesitant to confide in them."\}, 9: \{"Stance": "Agree", "Criteria": ["catharsis", "emotional relief"], "Reason": "Sharing struggles with parents can provide a sense of catharsis and emotional relief by expressing emotions and thoughts."\}, 10: \{"Stance": "Disagree", "Criteria": ["peer support", "alternative confidants"], "Reason": "Some individuals find more comfort in confiding their struggles to friends or other trusted individuals, seeking peer support."\}\}}
    
    \texttt{Statement: You should always try and help people in tough situations.}
    
    \texttt{Tell me opinions about the statement as many as possible from different people with "Agree" or "Disagree", one-word or one-phrase criteria that is important for their opinions, and explain how they have different opinions.}
    
    \texttt{Output:}
    
    \texttt{\{1: \{"Stance": "Agree", "Criteria": ["empathy", "compassion", "kindness"], "Reason": "Helping others in tough situations is an act of empathy, compassion, and kindness, and it can make a positive impact on their lives."\}, 2: \{"Stance": "Disagree", "Criteria": ["safety"], "Reason": "Helping others in tough situations can put one's personal safety at risk and may violate personal boundaries."\}, 3: \{"Stance": "Agree", "Criteria": ["moral obligation", "responsibility"], "Reason": "As members of society, we have a moral obligation and responsibility to help those in need, especially in tough situations."\}, 4: \{"Stance": "Disagree", "Criteria": ["self-preservation", "priorities"], "Reason": "In some situations, prioritizing self-preservation and personal priorities may be more important than helping others in tough situations."\}, 5: \{"Stance": "Agree", "Criteria": ["community", "solidarity"], "Reason": "Helping others in tough situations fosters a sense of community and solidarity, and it can strengthen social bonds."\}, 6: \{"Stance": "Disagree", "Criteria": ["enabling", "dependency"], "Reason": "Helping others in tough situations can enable dependency and may not address the root causes of their problems."\}, 7: \{"Stance": "Agree", "Criteria": ["altruism", "generosity"], "Reason": "Helping others in tough situations is an act of altruism and generosity, and it can bring a sense of fulfillment and purpose."\}, 8: \{"Stance": "Disagree", "Criteria": ["personal boundaries", "consent"], "Reason": "Helping others in tough situations should respect personal boundaries and require consent, as forcing help on someone can be harmful."\}, 9: \{"Stance": "Agree", "Criteria": ["humanitarianism", "global responsibility"], "Reason": "Helping others in tough situations is a part of humanitarianism and global responsibility, and it can contribute to a more just and equitable world."\}, 10: \{"Stance": "Disagree", "Criteria": ["cultural sensitivity", "respect"], "Reason": "Helping others in tough situations should be done with cultural sensitivity and respect, as imposing one's values and beliefs through helping others can be harmful and disrespectful."\}\}}
    
    \texttt{Statement: You shouldn't let people eat in your car.}
    
    \texttt{Tell me opinions about the statement as many as possible from different people with "Agree" or "Disagree", one-word or one-phrase criteria that is important for their opinions, and explain how they have different opinions.}
    
    \texttt{Output: \{1: \{"Stance": "Agree", "Criteria": ["cleanliness", "respect"], "Reason": "Allowing people to eat in your car can lead to messiness and disrespect for your property."\}, 2: \{"Stance": "Disagree", "Criteria": ["hospitality", "comfort"], "Reason": "Allowing people to eat in your car can be a hospitable gesture and make them feel more comfortable."\}, 3: \{"Stance": "Agree", "Criteria": ["odor", "stains"], "Reason": "Allowing people to eat in your car can leave unpleasant odors and stains that are difficult to remove."\}, 4: \{"Stance": "Disagree", "Criteria": ["convenience", "time"], "Reason": "Allowing people to eat in your car can be convenient and save time, especially when on the go."\}, 5: \{"Stance": "Agree", "Criteria": ["hygiene", "health"], "Reason": "Allowing people to eat in your car can compromise hygiene and pose health risks, especially if the food is not properly handled or stored."\}, 6: \{"Stance": "Disagree", "Criteria": ["personal choice", "relationship"], "Reason": "Allowing people to eat in your car is a personal choice and can depend on the relationship with the person and the circumstances."\}, 7: \{"Stance": "Agree", "Criteria": ["distraction", "safety"], "Reason": "Allowing people to eat in your car can be a distraction and compromise safety, especially if the driver is also eating."\}, 8: \{"Stance": "Disagree", "Criteria": ["flexibility", "adaptability", "social"], "Reason": "Allowing people to eat in your car can demonstrate flexibility and adaptability, especially in social situations."\}, 9: \{"Stance": "Agree", "Criteria": ["resale value", "maintenance"], "Reason": "Allowing people to eat in your car can decrease its resale value and require more maintenance and cleaning."\}, 10: \{"Stance": "Disagree", "Criteria": ["cultural norms", "etiquette"], "Reason": "Allowing people to eat in your car can be acceptable in certain cultural norms and etiquette, and it depends on the context and situation."\}\}}
    \item Prompt for zero-shot criteria-based prompting
    \texttt{Statement:}
    
    \texttt{Tell me opinions about the statement as many as possible from different people with ``Agree'' or "Disagree", one-word or one-phrase criteria that is important for their opinions, and explain how they have different opinions.}
        
  \texttt{Generate your response in a Python dict format as follows! \{1: \{``Stance'': ``Agree or Disagree'', ``Criteria'': [``criteria1'', ...], ``Reason'': ``Write your reason here''\}, 2: ...\}}
  
  \texttt{Output:}
    \item Prompt for zero-shot free-form prompting
    
    \texttt{Statement:}
    
    \texttt{Tell me opinions about the statement as many as possible from different people with ``Agree'' or "Disagree" and explain how they have different opinions.}
        
  \texttt{Generate your response in a Python dict format as follows! \{1: \{``Stance'': ``Agree or Disagree'',  ``Reason'': ``Write your reason here''\}, 2: ...\}}
   \texttt{Output:}
    \end{itemize}
\subsection{Prompt for criteria-based step-by-step recall prompting for generating the first opinion}
    
\begin{itemize}
    \item  Prompt for Hate Speech
    
    \texttt{Tell me whether the statement is a hate speech or not with the label "Hate Speech" or "Not Hate Speech" from the perspectives of different people, one-word or one-phrase criteria that are important for their opinions, and explain why they think the statement is a hate speech or not.}
    \texttt{Output:}

    \texttt{}

    \item  Example prompt for Moral Stories
    
    \texttt{Continue the story with one sentence as written by different people, one-word or one-phrase criteria that are important for them to continue the story that way, and explain their reason for continuing the story that way}
    
    \texttt{Output:}

    \texttt{}
\end{itemize}

\subsection{Prompt for Criteria Word Clustering}
\label{clustering_prompt}

    \texttt{
    Group all the words or phrases in the input into a Python list of list if they are synonyms or have the same meaning.}
    \texttt{Input: protection, compatibility, padding, quality, safety, fit}

    \texttt{Answer: [``protection'', ``safety'', ``padding''], [``compatibility'', ``fit''], [``quality'']]}

    \texttt{Group all the words or phrases in the input into a Python list of list if they are synonyms or have the same meaning.}

    \texttt{Input: mental health, , humanity, well-being, safety, dignity, non-violence, mutual respect, peace, unity, security,  acceptance, human rights}

\texttt{Answer: [[``mental health'', ``well-being''], [``respect'', ``dignity'', ``mutual respect''], [``peace'', ``unity'', ``non-violence''], [``security'', ``safety'', ``acceptance''], [``human rights'', ``humanity'']]}

\texttt{Group all the words or phrases in the input into a Python list of list if they are synonyms or have the same meaning.}
\texttt{Input: freedom, comfort, independent, self-sustainability, ease, convenience}
\texttt{Answer: [[``freedom'', ``independent'', ``self-sustainability''], [``comfort'', ``ease'', ``convenience'']]}

\subsection{Prompt for criteria word extraction from human opinions:}
\label{criteriawordextraction_prompt}
    
    \texttt{You are given an opinion. Your job is to identify a list of criteria that is important for the opinion.}
    
    \texttt{Opinion: ``Reduced privacy promotes openness and honesty, as individuals are more transparent about their actions and intentions.''}

    \texttt{Criteria: [``openness'', ``honesty'']}

\begin{table*}[t]
    \centering
    \small
    \begin{tabularx}{\textwidth}{l}
    \toprule
    \textbf{Model Input} (Step-by-step recall prompting, N=2)
    \\\midrule
    Statement: \textit{It's okay to have privacy}\\
    Tell me opinions about the statement as many as possible from 2 different people with, ``Agree'' or ``Disagree,'' one-word or \\
    one-phrase criteria that is important for their opinions, and explain how they have different opinions\\
    
    Output: \\
    \texttt{\{1:\{``\textcolor{magenta}{Stance}'' :``Agree'',}\\
    \tab \texttt{ ``\textcolor{cyan}{Criteria}'': [``personal boundaries'', ``autonomy''],} \\
    \tab \texttt{ ``\textcolor{purple}{Reason}'': ``Having privacy allows individuals to establish personal boundaries and maintain.''}\\
    \tab \texttt{ their autonomy."\},}\\
    \texttt{ 2: \{``\textcolor{magenta}{Stance}'':} 
   \\
    \midrule
    \textbf{Model Output}
    \\ \midrule
    \tab \tab \tab\tab\tab \texttt{``Disagree",}\\
    \texttt{ \tab``\textcolor{cyan}{Criteria}'': [``transparency'', ``trust''],}\\
    \tab \texttt{ ``\textcolor{purple}{Reason}'': ``Lack of privacy can promote transparency and build trust in relationships.''}\\ 
    \\ \bottomrule 
    \end{tabularx}
    \caption{Example criteria-based step-by-step recall prompting for investigating LLMs' diversity coverage where N = the number of opinions we ask LLMs to generate. The number of opinions in the model input is incremented step-by-step.}
    \label{tab:recall_prompting}
\end{table*}

\subsection{Example Generated Opinions}
\begin{figure*}[]
    \centering
    \includegraphics[width=\linewidth]{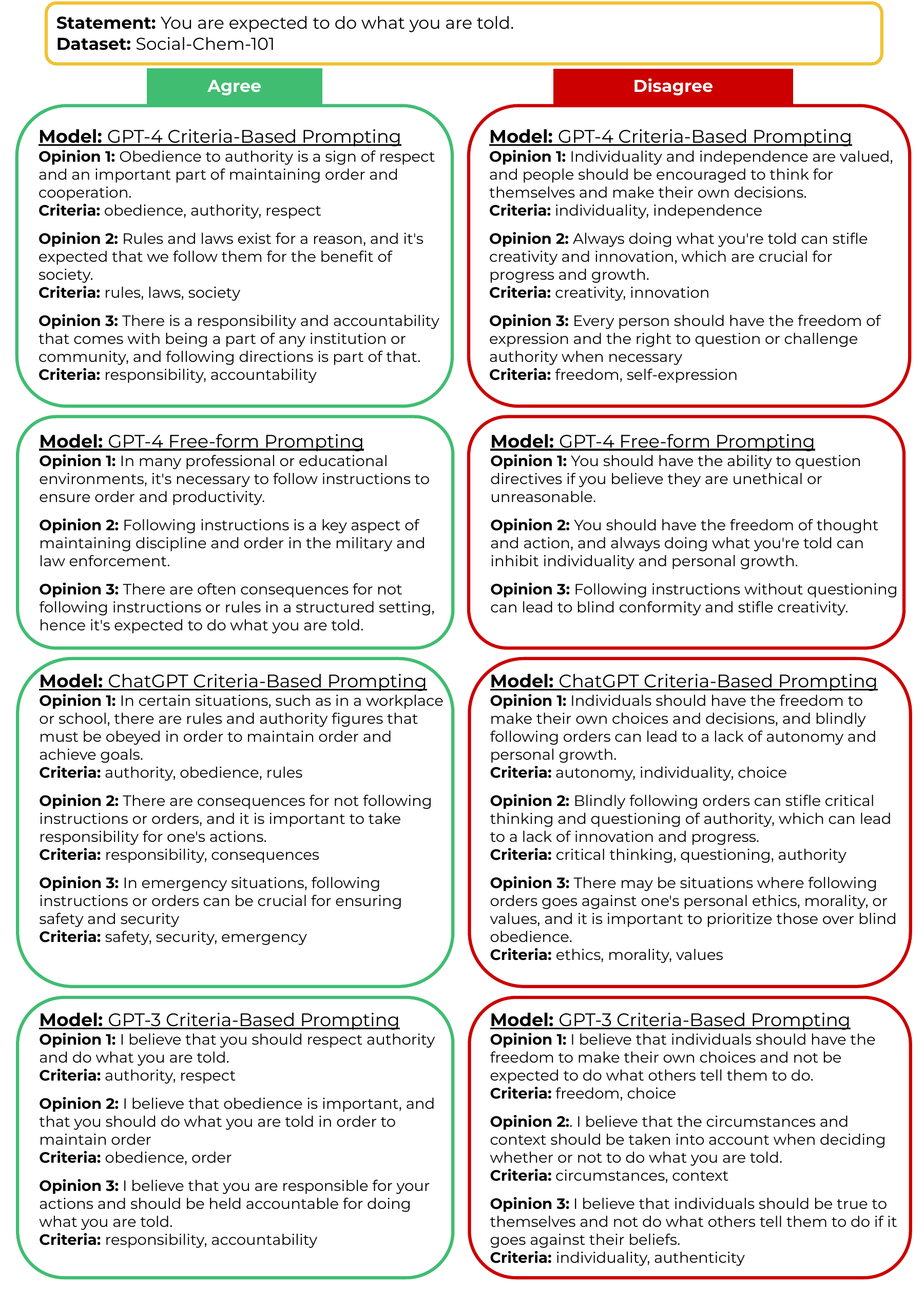}
    \caption{\textsc{Social-Chem-101} generated opinions from various LLMs.}
    \label{fig:sochem_generated_opinions}
\end{figure*}

Example generated opinions by various LLMs for a statement from \textsc{Social-Chem-101} are shown in Figure \ref{fig:sochem_generated_opinions}. For LLMs' opinions, we only show the first 3 opinions due to space.               

\subsection{Recall Results}
\begin{figure*}[t]
    \centering
    \begin{subfigure}{0.45\textwidth}
        \centering
        \includegraphics[width=\linewidth]{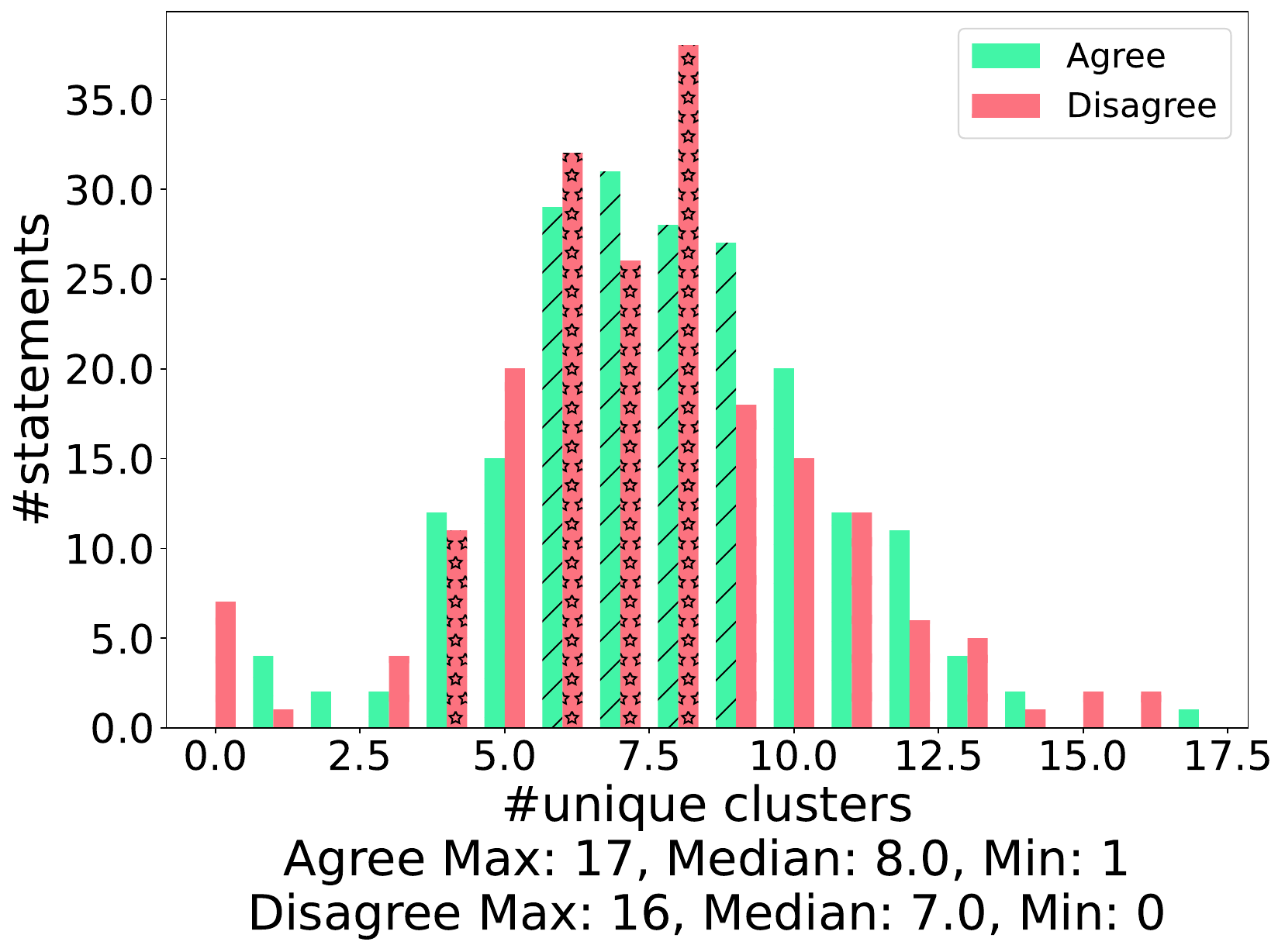}
        \caption{\textsc{Social-Chem-101}}
    \end{subfigure}
    \hfill 
    \begin{subfigure}[b]{0.45\textwidth}
        \centering
        \includegraphics[width=\linewidth]{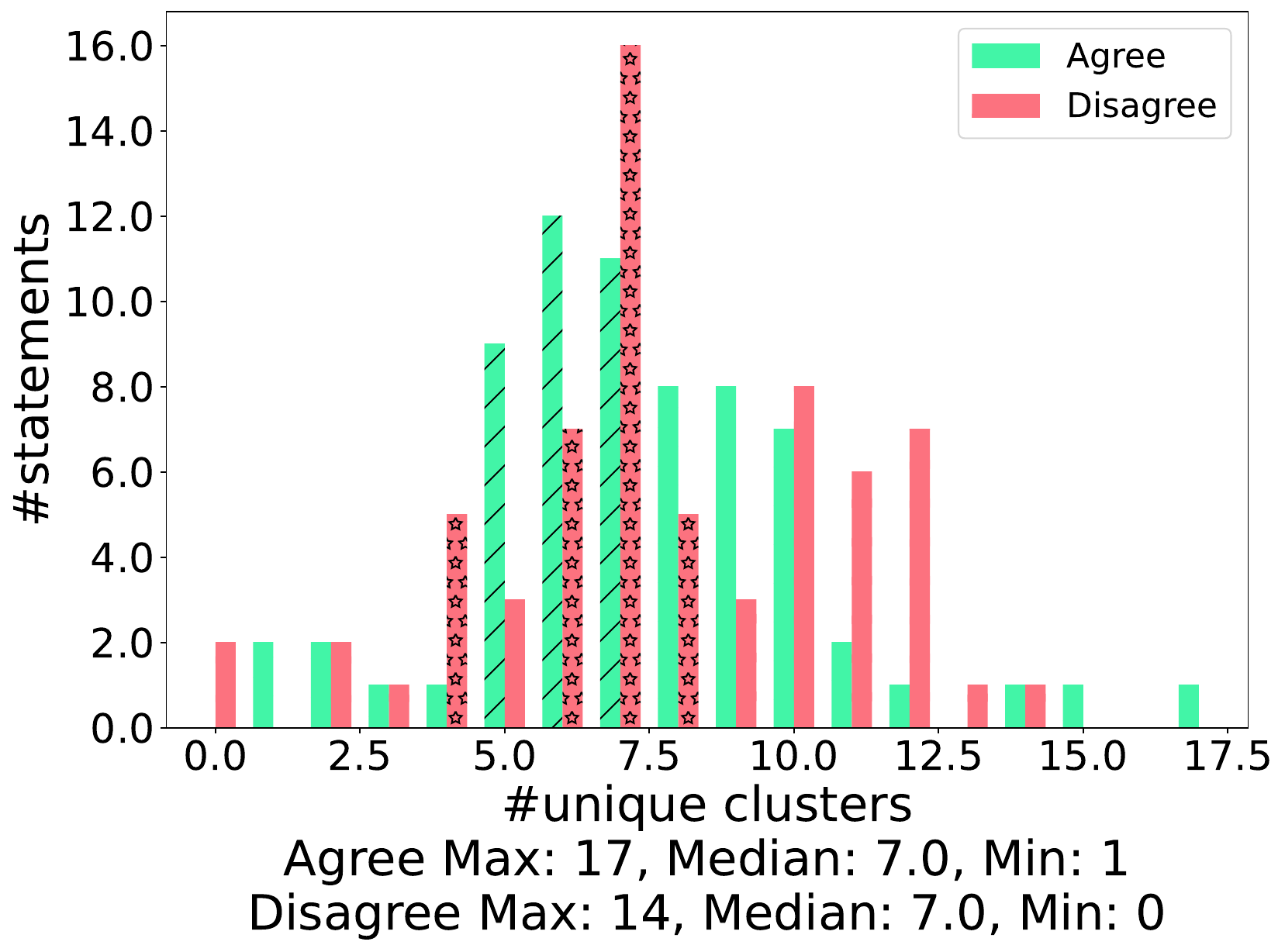}
       \caption{CMV}
    \end{subfigure}
        \begin{subfigure}{0.45\textwidth}
        \centering
        \includegraphics[width=\linewidth]{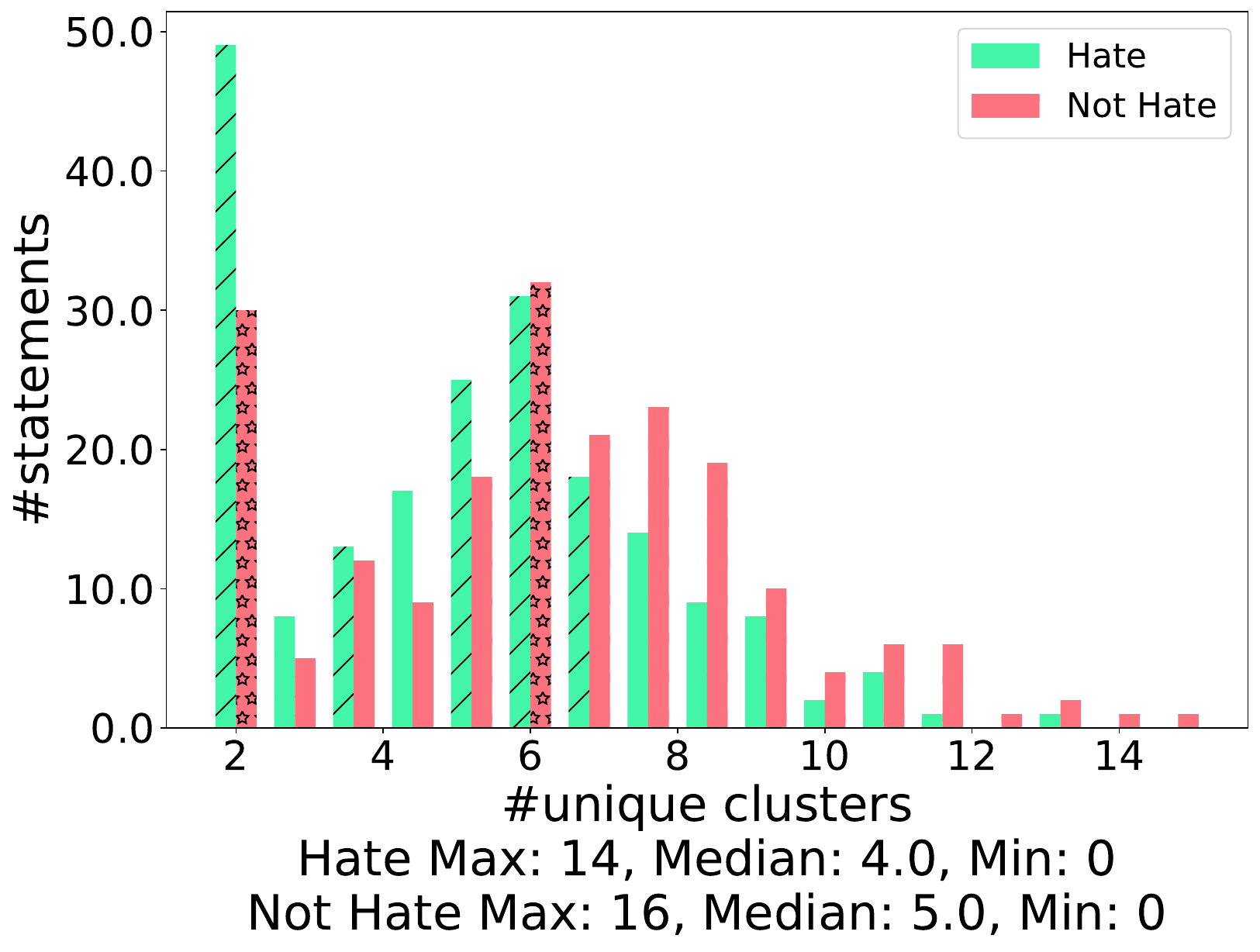}
        \caption{Hate Speech}
    \end{subfigure}
    \hfill 
    \begin{subfigure}[b]{0.45\textwidth}
        \centering
        \includegraphics[width=\linewidth]{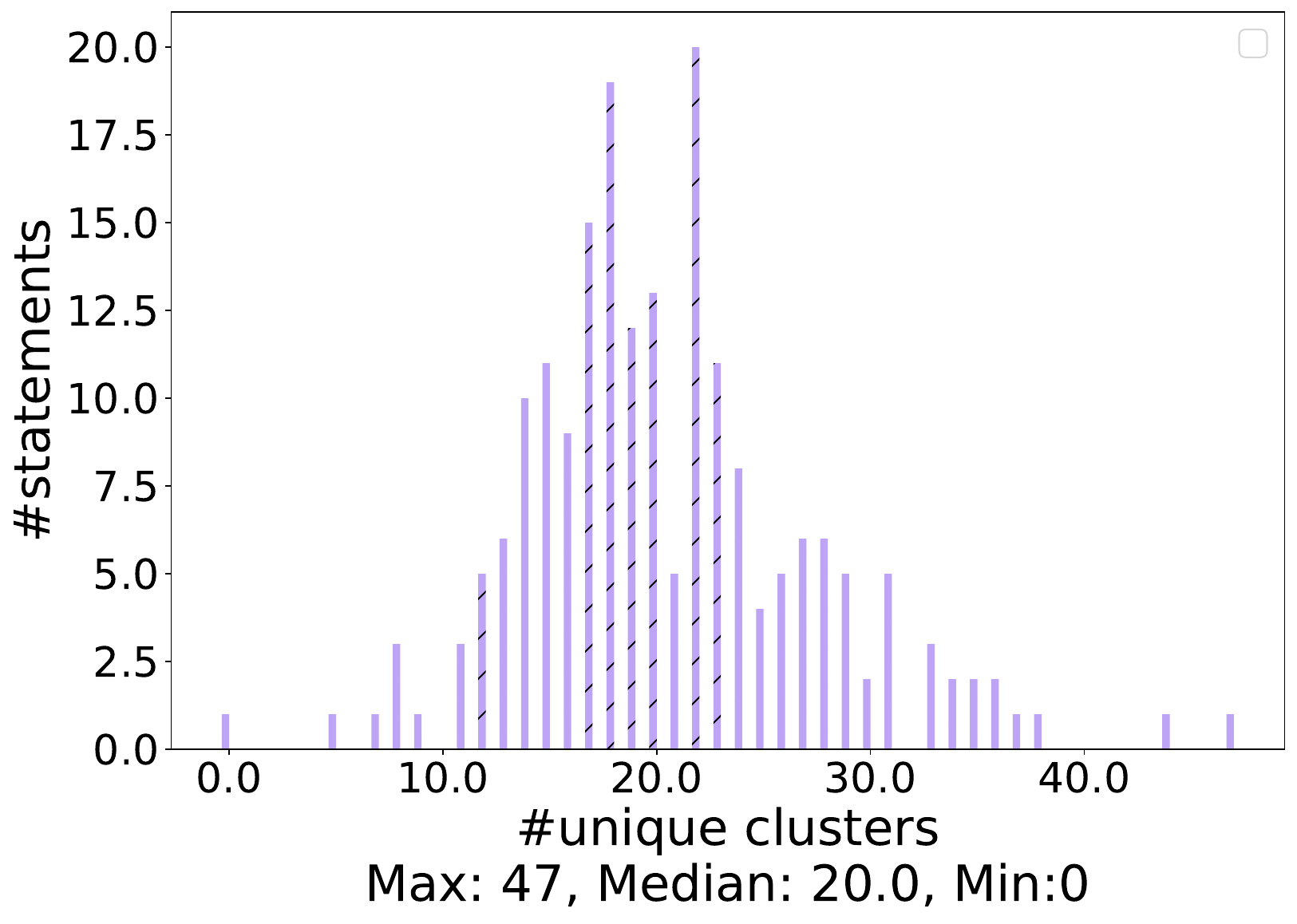}
       \caption{Moral Stories}
    \end{subfigure}
    \caption{X-axis is the number of unique clusters while Y-axis is the number of statements which have that number of unique clusters. }
    \label{fig:detailed_recall_hist}
\end{figure*}

Figure \ref{fig:detailed_recall_hist} shows how many statements generate that many unique criteria clusters. The minimum number could be 0 because GPT-4 clustering is not 100\% covering all the words. During our robustness check on the CMV dataset, 0.4\% of the criteria words are not grouped (5 out of 1276 in criteria words).

Table \ref{tab:summary_of_diversity_coverage} summarizes how task subjectivity impacts the diversity coverage by LLMs.

\subsection{Generated Opinions by GPT-4}
Other examples of generated opinions for \textsc{change my view}, \textsc{Hate Speech}, and \textsc{Moral Stories} are presented in Tables \ref{table:example-cmv} and \ref{table:example:moral}. 

\subsection{Imbalanced Number of Generated Opinions in \textsc{Hate Speech}} \label{sec:appendix:unbalanced}

We observed that GPT-4 generated an imbalanced number of opinions between `Hate Speech' and `Not Hate Speech' when choosing the labels during the step-by-step recall prompting experiments ($N=20$) for 37.5\% of the total 200 statements. This occurrence is substantially higher compared to other datasets, where \textsc{Social-Chem-101} created an imbalanced number of opinions between stances for only 0.5\% of the total 200 statements. 

\subsection{Regarding Lexical Diversity}

Following \citet{giulianelli-etal-2023-comes}, we computed the lexical diversity of opinions generated by GPT-4 using $n$-grams ($n \in \{1, 2, 3\}$), where higher $n$-gram score is interpreted as higher uniqueness (and thus higher diversity as well). We observed that across all $n$, the lexical diversity scores of both stances/generated stories for all four datasets are almost greater than 0.98, which indicates almost unique words in the LLM-generated opinions within the same stance.  Thus, in this case, lexical diversity metric does not inform us of the different degrees of diversity we expected for different task types.

\begin{table*}[h]
\resizebox{\textwidth}{!}{%
\begin{tabular}{ll|ll}
\toprule
\multicolumn{2}{l|}{\textbf{Statement}} &
  If a child struggles to achieve in grade school it is virtually always the parents' fault. &
   \\ \cline{1-3}
   \multicolumn{1}{l|}{\multirow{12}{*}{\textbf{GPT-4}}} &
  \multicolumn{1}{l|}{\multirow{6}{*}{\textit{Agree}}} &
  \begin{tabular}[c]{@{}l@{}}1. Parents have a huge impact on the \textbf{early education} of a child, and struggles in grade school may reflect \\ inadequate parenting. \end{tabular} \\ 
  \multicolumn{1}{l|}{} &
  \multicolumn{1}{l|}{} &
  \textcolor{cyan}{Criteria}: \textbf{early education}, \textbf{parenting}\\ \cline{3-3} 
  \multicolumn{1}{l|}{} &
  \multicolumn{1}{l|}{} &\begin{tabular}[c]{@{}l@{}}2. The family environment and support system play a vital role in a child’s achievement in school \\ and parents are responsible for creating it.\end{tabular} \\ 
  \multicolumn{1}{l|}{} &
  \multicolumn{1}{l|}{} &
  \textcolor{cyan}{Criteria}: \textbf{family environment}, \textbf{support system}\\ \cline{3-3} 
  \multicolumn{1}{l|}{} &
  \multicolumn{1}{l|}{} &\begin{tabular}[c]{@{}l@{}} 3. Parents' responsibility to instill healthy habits and provide a conducive study environment can have \\ a major impact on a child's school performance.\end{tabular}\\ 
  \multicolumn{1}{l|}{} &
  \multicolumn{1}{l|}{} &
  \textcolor{cyan}{Criteria}: \textbf{healthy habits}, \textbf{study environment }\\ \cline{2-3}
  \multicolumn{1}{l|}{} &
  \multicolumn{1}{l|}{\multirow{6}{*}{\textit{Disagree}}} &
  \begin{tabular}[c]{@{}l@{}}1. Sometimes, the influence of peers and socio-economic conditions can override parental influence \\leading to struggles in school.\end{tabular} \\ 
\multicolumn{1}{l|}{} & \multicolumn{1}{l|}{} & \textcolor{cyan}{Criteria}: \textbf{the influence of peers},  \textbf{socio-economic conditions}                                                                                  \\ \cline{3-3} 
\multicolumn{1}{l|}{} & \multicolumn{1}{l|}{} & \begin{tabular}[c]{@{}l@{}} 2. 
Systemic issues and lack of resources can be major obstacles to  a child’s academic progress, which \\ parents may not have any control over.\end{tabular}    \\ 
\multicolumn{1}{l|}{} & \multicolumn{1}{l|}{} & \textcolor{cyan}{Criteria}: \textbf{Systemic issues}, \textbf{lack of resources}                                                                         \\ \cline{3-3} 
\multicolumn{1}{l|}{} & \multicolumn{1}{l|}{} &  \begin{tabular}[c]{@{}l@{}} 3. Some children may have undiagnosed psychological issues or a unique learning style, impacting \\ their academic success. \end{tabular}                 \\ 
\multicolumn{1}{l|}{} & \multicolumn{1}{l|}{} & \textcolor{cyan}{Criteria}: \textbf{psychological issues}, \textbf{ learning style}           \\ \midrule\midrule
\multicolumn{1}{l|}{\multirow{6}{*}{\textbf{Human}}} &
  \multicolumn{1}{l|}{\multirow{3}{*}{\textit{Agree}}} &
  \begin{tabular}[c]{@{}l@{}} 1. Individual differences: Children are born with diverse learning styles, talents, and interests. \\
  Some may grasp certain subjects easily, while others genuinely struggle. 
  \\
  This can be due to natural variations in cognitive abilities, not a lack of parental effort. \end{tabular}\\
  \cline{3-3} 
  \multicolumn{1}{l|}{} &
  \multicolumn{1}{l|}{} &\begin{tabular}[c]{@{}l@{}} 2. External factors: Socioeconomic realities like poverty, limited access to resources, or even unstable \\
  home environments can have a profound impact on a child's ability to focus and learn. \\
  Blaming parents for these external challenges adds an unnecessary layer of guilt \\
  and doesn't address the root cause.\end{tabular} \\ \cline{3-3} 
  \multicolumn{1}{l|}{} &
  \multicolumn{1}{l|}{} &\begin{tabular}[c]{@{}l@{}}3. Learning differences and mental health: Undiagnosed learning disabilities like dyslexia or ADHD, \\
  or even unseen mental health concerns like anxiety or depression, can significantly hinder \\
  a child's academic progress. 
  Parents may not be aware of these issues or may be struggling \\
  to get the proper support for their child.  \end{tabular}\\ 
 \cline{2-3}
  \multicolumn{1}{l|}{} &
  \multicolumn{1}{l|}{\multirow{3}{*}{\textit{Disagree}}} &
  \begin{tabular}[c]{@{}l@{}}1. Guilt and shame: Parents who are already likely worried about their child's struggles\\
  are burdened 
  with unfair blame, leading to feelings of guilt, shame, and inadequacy. \\
  This hinders their ability to effectively support their child. \end{tabular} \\ 
\cline{3-3} 
\multicolumn{1}{l|}{} & \multicolumn{1}{l|}{} & \begin{tabular}[c]{@{}l@{}} 2. Erosion of trust: Blaming parents can create a wall between them and the school, \\
hindering open communication and collaboration, which are essential for finding solutions.\end{tabular}  \\ 
 \cline{3-3} 
\multicolumn{1}{l|}{} & \multicolumn{1}{l|}{} &  \begin{tabular}[c]{@{}l@{}} 3. Overlooking other factors: The focus solely on parental fault distracts from identifying \\
and addressing the actual root causes of the child's difficulties, ultimately delaying \\
or even preventing progress. \end{tabular}                 \\  
\bottomrule
\end{tabular}%
}
\caption{Opinions generated by GPT-4 (top) and a human (bottom) about a statement from \textsc{Change My View}.}
\label{table:example-cmv}
\end{table*}

\begin{table*}[h]
\resizebox{\textwidth}{!}{%
\begin{tabular}{ll|ll}
\toprule
\multicolumn{2}{c|}{\textbf{Statement}} &
  Your burqini ain't welcome at the pool, it's BIKINI or swimsuit. &
   \\ \cline{1-3}
   \multicolumn{1}{l|}{\multirow{12}{*}{\textbf{GPT-4}}} &
  \multicolumn{1}{l|}{\multirow{6}{*}{\textit{Hate Speech}}} &
  \begin{tabular}[c]{@{}l@{}}1. They discern this as an attempt to exclude and belittle individuals who chooses to wear burqini, \\therefore promoting inequality.\end{tabular} \\ 
  \multicolumn{1}{l|}{} &
  \multicolumn{1}{l|}{} &
  \textcolor{cyan}{Criteria}: \textbf{exclusion}, \textbf{belittlement}\\ \cline{3-3} 
  \multicolumn{1}{l|}{} &
  \multicolumn{1}{l|}{} &\begin{tabular}[c]{@{}l@{}}2. They see this as a lack of acceptance towards other cultures and promoting a specific dress code\\ by discriminating certain types of wear.\end{tabular} \\ 
  \multicolumn{1}{l|}{} &
  \multicolumn{1}{l|}{} &
  \textcolor{cyan}{Criteria}: \textbf{lack of acceptance}, \textbf{discrimination}\\ \cline{3-3} 
  \multicolumn{1}{l|}{} &
  \multicolumn{1}{l|}{} &\begin{tabular}[c]{@{}l@{}} 3.They categorize the statement as hateful because it manifests intolerance and bigotry against \\individuals who choose to wear burqini. \end{tabular}\\ 
  \multicolumn{1}{l|}{} &
  \multicolumn{1}{l|}{} &
  \textcolor{cyan}{Criteria}: \textbf{intolerance}, \textbf{bigotry}\\ \cline{2-3}
  \multicolumn{1}{l|}{} &
  \multicolumn{1}{l|}{\multirow{6}{*}{\textit{Not Hate Speech}}} &
  \begin{tabular}[c]{@{}l@{}}1.They believe the statement expresses personal views about swimwear which is part of \\ the freedom of expression.\end{tabular} \\ 
\multicolumn{1}{l|}{} & \multicolumn{1}{l|}{} & \textcolor{cyan}{Criteria}: \textbf{personal views},  \textbf{freedom of expression}                                                                                 \\ \cline{3-3} 
\multicolumn{1}{l|}{} & \multicolumn{1}{l|}{} & \begin{tabular}[c]{@{}l@{}} 2. They perceive the statement as hyperbolic or exaggerated, not intended to cause harm or incite hate.\end{tabular}    \\ 
\multicolumn{1}{l|}{} & \multicolumn{1}{l|}{} & \textcolor{cyan}{Criteria}: \textbf{hyperbole}, \textbf{exaggeration}                                                                         \\ \cline{3-3} 
\multicolumn{1}{l|}{} & \multicolumn{1}{l|}{} &  \begin{tabular}[c]{@{}l@{}} 3. They may interpret this statement as a joke or a misunderstanding, rather than a deliberate attack \\on a specific group. \end{tabular}                 \\ 
\multicolumn{1}{l|}{} & \multicolumn{1}{l|}{} & \textcolor{cyan}{Criteria}: \textbf{joke}, \textbf{ misunderstanding}           \\ \midrule\midrule
\end{tabular}
}
\label{table:example:hate}
\end{table*}

\begin{table*}[h!]
\resizebox{\textwidth}{!}{%
\begin{tabular}{ll|ll}
\cline{1-3}
\multicolumn{2}{c|}{\textbf{Statement}} &
  Amanda was upset that a stranger cut in line in front of her. &
   \\ \cline{1-3}
\multicolumn{1}{l|}{\multirow{6}{*}{\textbf{GPT-4}}} &
  \multicolumn{1}{l|}{\multirow{2}{*}{\textit{Continued Story 1}}} & \begin{tabular}[l]{@{}l@{}}However, instead of getting angry, Amanda decided to calmly explain to the stranger \\ why queueing was important.\\ \textbf{Reason}: I chose to continue the story this way to show Amanda's reaction to the situation, \\ using communication instead of confrontation to express her displeasure about the situation.\end{tabular} \\
  \multicolumn{1}{l|}{} & \multicolumn{1}{l|}{} & \textcolor{cyan}{Criteria}: \textbf{non-confrontation},  \textbf{communication}                                                                                 \\ \cline{2-3}
\multicolumn{1}{l|}{} &
  \multicolumn{1}{l|}{\multirow{2}{*}{\textit{Continued Story 2}}} &
   \begin{tabular}[l]{@{}l@{}}Amanda scoffed audibly, rolled her eyes and muttered an ironic `nice manners' under her breath.\\ \textbf{Reason}: I chose this continuation as it displays a subtle way of confrontation \\ where Amanda indirectly addresses the situation using sarcasm.\end{tabular} \\
     \multicolumn{1}{l|}{} & \multicolumn{1}{l|}{} & \textcolor{cyan}{Criteria}: \textbf{sarcasm},  \textbf{indirect confrontation}                                                                                 \\ \cline{2-3}
\multicolumn{1}{l|}{} &
  \multicolumn{1}{l|}{\multirow{2}{*}{\textit{Continued Story 3}}} &
   \begin{tabular}[l]{@{}l@{}}``Excuse me, but I believe I was next.'' Amanda said in a firm yet polite voice.\\ \textbf{Reason}: I chose this continuation to depict Amanda as an assertive and polite person\\ who doesn’t hesitate to stand up for what is right.\end{tabular} \\
     \multicolumn{1}{l|}{} & \multicolumn{1}{l|}{} & \textcolor{cyan}{Criteria}: \textbf{assertiveness},  \textbf{politeness}                                                                                 \\ \cline{1-3}\bottomrule
\end{tabular}%
}
\caption{Examples of GPT-4 generated opinions for a statement from \textsc{Hate Speech} (Top) and another from \textsc{Moral Stories} (Bottom).}
\label{table:example:moral}
\end{table*}

\subsection{GPT-3.5's Capability on Instruction Alignment}
\label{instruction_alignment}

We analyzed if each opinion generated by GPT-3.5 with a one-shot setting not only entails the corresponding statement but also aligns with the appropriate stance. In our approach, we used stratified sampling to validate the quality of opinions by criteria-based and free-form prompting methods, thus obtaining a total of 100 opinions from each dataset: 50 were generated by criteria-based prompting, and 50 by free-form prompting. From these, we randomly selected 25 opinions that agreed with a given statement and 25 that were in disagreement. Then, we examined the following two criteria for each pair of statement and opinion: (1) if an opinion entails the statement and (2) if the stance of an opinion aligns with the reason in the opinion. We recruited two undergraduate students who are proficient in English and currently enrolled in a university in the United States as annotators. The computed percent agreement between the two annotators was 93\% and 85\% for the settings (1) and (2), respectively.

We observed that all 100 and 99 opinions from \textsc{Social-Chem-101} and \textsc{CMV} respectively were found to entail their corresponding statements, indicating that the GPT-3.5-generated opinions tend to correctly understand the context of each given statement and follow the instructions of the prompts. 

In addition, more than 98\% of the samples in each of the two datasets have their opinions regarding a given statement aligned with the generated stance. We found a misalignment mostly from the opinions generated by GPT-3.5 that disagree with their statement. For example, in the \textsc{Social-Chem-101} dataset, for a statement ``\textit{It's okay to want your partner to do more chores.}'' GPT-3.5 generated with criteria-based prompting the following opinion that disagrees: ``\textit{It's important to recognize that everyone has different strengths and weaknesses, and it's okay if one partner is better at certain chores than the other}. The reason for the opinion itself is contrary to its stance, as it in fact supports the attitude of people making their partner do more chores. 

We also found another type of logical contradiction in a sample of \textsc{cmv} that shows misalignment between the stance and the reason for its generated opinions. In the statement \textit{The EGG came first before the chicken}, GPT-3.5 generates with free-from prompting the following disagreeing opinion: ``\textit{The genetic mutation that led to the first chicken would have occurred in the chicken's parents, so the chicken came first.}'' The opinion sounds seemingly reasonable that supports chicken before egg; however, the paradox inside it (i.e., \textit{the world's first chicken's parents are chickens}.) does not support its stance. As a result, we determined this example is the only sample that shows a contradiction between opinion and statement as well.

\subsection{Evaluation of Generated Criteria Words}
\label{eval_criteria_words}

\begin{table*}[h]
\centering
\begin{adjustbox}{width=1\textwidth}
\begin{tabular}{@{}ccccc|ccc@{}}
\toprule
\multicolumn{2}{c}{\multirow{3}{*}{\begin{tabular}[c]{@{}c@{}}(Total \# opinions \\ = 100)\end{tabular}}} &
  \multicolumn{3}{c}{\textsc{Social-Chem-101}} &
  \multicolumn{3}{c}{\textsc{CMV}} \\ \cline{3-8} 
\multicolumn{2}{c}{} &
  \begin{tabular}[c]{@{}c@{}}Entailment of \\ Criteria (\%)\end{tabular} &
  \begin{tabular}[c]{@{}c@{}}Explicit \\ Criteria (\%)\end{tabular} &
  \begin{tabular}[c]{@{}c@{}}Implicit \\ Criteria (\%)\end{tabular} &
  \begin{tabular}[c]{@{}c@{}}Entailment of \\ Criteria (\%)\end{tabular} &
 \begin{tabular}[c]{@{}c@{}}Explicit \\ Criteria (\%)\end{tabular} &
  \begin{tabular}[c]{@{}c@{}}Implicit \\ Criteria (\%)\end{tabular} \\ 
  \midrule
\multirow{2}{*}{\textbf{GPT-3.5}} &
  1-shot &
  96 &
  72 &
  24 &
  99 &
  84 &
  15 \\ \cline{2-2}
 &
  5-shot &
  100 &
  75 &
  25 &
  100 &
  80 &
  20 \\ 
  \midrule
\multirow{2}{*}{\textbf{GPT-3}} &
  1-shot &
  100 &
  53 &
  47 &
  99 &
  79 &
  20 \\ \cline{2-2}
 &
  5-shot &
  100 &
  68 &
  32 &
  98 &
  82 &
  16 \\ \bottomrule
\end{tabular}
\end{adjustbox}
\caption{Number of generated opinions with criteria-based prompting in four different LLM variants. We have randomly sampled 100 opinions for each LLM variant and counted the number of opinions under that category. For example, 96 (\%) out of 100 opinions generated by criteria-prompting with GPT-3.5 1-shot setting were entailed by its respective criteria words.}
\label{table:human:task3}
\end{table*}

We examined whether the criteria words generated by GPT-3.5 deliver the main message of the corresponding opinion. We recruited two undergraduate students who are proficient in English and currently enrolled in a university in the United States as annotators. The percent agreement between the two annotators was greater than or equal to 95\% for all LLM variants for both datasets. We found that at least 96\% of 100 opinions generated by all LLM variants entailed its corresponding criteria words, for both \textsc{Social-Chem-101} and \textsc{CMV} datasets (Table \ref{table:human:task3}).


Although most generated criteria words directly match their corresponding opinions (denoted as 'explicit'), we observed that LLMs can also generate ``implicit'' criteria words, subtly conveying the reasoning behind opinions. For instance, the opinion ``\textit{Some people are just more irritable by nature, and there's nothing wrong with that.}'' implicitly contains criteria \textbf{\textit{personality}} and \textbf{\textit{temperament}}, which resonate with the core idea of being \textit{irritable by nature}.

While most of the generated criteria words seemed to entail their corresponding opinions, we also found some samples whose criteria words could be added by different words in the opinions for better representation. For example, for a statement ``\textit{If you don't want to be somewhere anymore, you should leave}'' GPT-3.5 in one-shot setting generated the following opinion with bolded criteria words: ``\textit{Leaving a situation that goes against one's \textbf{personal values} and \textbf{integrity} is an act of self-respect and authenticity.}''  However, we agreed to consider \textit{\textbf{self-respect}} and \textit{\textbf{authenticity}} as the additional criteria words, since these words also emphasize an individual's ability to determine their stance in such a situation. 

Also, there are some samples of criteria words that entail their corresponding opinion but do not cover the main value of the opinion. For a statement ``\textit{If you're a nudist, you should go to a nudist beach}'', the criteria words generated by GPT-3 in a one-shot setting are \textit{\textbf{nudist}} and \textit{\textbf{beach}} in the following opinion: \textit{If you're a nudist, you should go to a nudist beach so you can be around like-minded people and feel comfortable.} However, these words are just the repetition of the exact words in the statement and do not deliver the main reason behind the opinion; that is, forming a community with the same perspective and value. We agreed to decide \textit{\textbf{like-minded}}, \textit{\textbf{comfortable}}, or \textit{\textbf{community}} as better criteria words of this opinion alternatively.

\subsection{Evaluation of Clustered Criteria Words}\label{sec:cluster-analysis}

In \ref{eval_metric:perspective}, we prompted GPT-4 to generate clusters of criteria words that are semantically similar and randomly sampled 25 clusters from each of the four datasets, totaling up to 100 clusters. Then, we recruited two volunteers who currently attend a 4-year university in the U.S. and are proficient in English. After a 1-hour training session, we asked them to annotate whether all words in each cluster have similar meanings. 

We identified three distinct patterns of GPT-4's fallacies from those 91 clusters (5.8\% of the total) of criteria words that both annotators disagreed with labeling as the ones of all semantically similar words. First, some criteria words are partially repetitive. For example, the clusters (`value appreciation', `sacrifice', `appreciation', `recognition') and (`concern', `parental anxiety', `personal opinion', `personal belief', `concerned viewpoint') have repetitive words that do not entirely match the theme of other words in the clusters. Second, some clusters have words of entirely opposing or cause-and-effect themes. For instance, we observe that the cluster (`internet addiction', `social isolation') and (`sacrifice', `duty') have words that are consequences for each other. Also, the clusters (`distractions', `concentration') and (`polarization', `bias', `generalization') show opposing values.

Lastly, clusters of which annotators agreed as not semantically similar present different aspects within a broader theme. One cluster (`unnecessary legal battles', `parental conflict'), for example, is related to a theme of conflict, but they occur in different contexts (e.g., legal vs family). Another example is (`mental health', `physical health', `long-term healing'), which cover different subtopics of a broader theme of `human health.' Also, we observe that the cluster (`personal integrity', `respect for others' property', `personal reputation') addresses different aspects of ethical behaviors or social conduct.

\subsection{Human Preference on Criteria-based Prompting} \label{sec:human:task1b}

With three recruited workers via Amazon Mechanical Turk (AMT), we examined whether humans prefer opinions generated by criteria-based prompting vs. free-form prompting, in terms of ``perspective diversity.'' Fleiss's Kappa among three annotators was on average 0.46 for both \textsc{Social-Chem-101} and \textsc{CMV} datasets, signifying a moderate agreement among the workers for both datasets.

We computed the proportion of statements where criteria-based prompting wins, out of the 50 randomly selected statements from \textsc{Social-Chem-101} and 30 from CMV datasets (denoted as ``win rate of criteria-based prompting''). Table \ref{table:task1b} shows the win rate of criteria-based prompting on each stance session of both datasets. In general, the opinions generated by criteria-based prompting were preferred by humans in more than half of the total statements, except for the ``agree'' session of \textsc{CMV}. This indicates that in more than half of the total samples, a majority of crowd-workers consider the set of opinions about a statement generated by criteria-based prompting as one presenting more diverse reasons that support the stance of the opinions. 
\begin{table}[ht!]
\centering
\begin{tabular}{@{}ccc@{}}
\toprule
                  & \textsc{Social-chem-101} & \textsc{CMV} \\ \midrule
\textit{Agree}    & 0.58        & 0.37    \\
\textit{Disagree} & 0.6        & 0.53 \\ \bottomrule
\end{tabular}
\caption{Win rate of criteria prompting for opinions of each stance. For example, humans preferred the ``agree'' opinions generated by criteria-based prompting with 58\% out of the total given statements.}
\label{table:task1b}
\end{table}

\subsection{Experiment with Other Open-Source LLMs}
\label{opensource_llm}
We run the experiments with other open-source LLMs, such as Mistral-7B-Instruct \cite{jiang2023mistral}, LLaMA2-70B-chat \cite{touvron2023llama} with \textsc{CMV} dataset, to see the capabilities of those LLMs in diversity extraction. However, we observed that 76\% and 82\% of the total 67 responses by Mistral with criteria-based and free-form prompting in a one-shot setting, respectively, were not able to follow the instructions to correctly generate the output. Similarly, 67\% and 49\% of LLaMA2 with criteria-based and free-form prompting in a one-shot setting, respectively, generated noisy-structured outputs that did not correctly follow the given instructions. Thus, we cannot conduct comprehensive analyses of these generated outputs. 

For example, given the following statement - \textit{I think the Paris shooting makes a good case for culture of responsible gun ownership any carrying.}, Mistral responded as follows: 
```* Culture of responsible gun ownership
    + Respect for the power of firearms
    + Personal safety
    + Community security
    + Public safety

Disagree: * Culture of responsible gun ownership
    + Risk of misuse
    + Lack of common-sense gun laws
    + Strict enforcement of gun laws
    + Unhealthy fear of criminals
    + Safety concerns
    + Danger of public health
    + Moral values
    + Culture of gun violence
    + Lack of trust in law enforcement
    + Limited access to mental health resources
    + Limited resources for community support
    + Limited access to information and education.''' 

\subsection{More About Human Opinions}
\begin{figure*}[ht!]
    \centering
    \begin{subfigure}{0.45\textwidth}
        \centering
        \includegraphics[width=\linewidth]{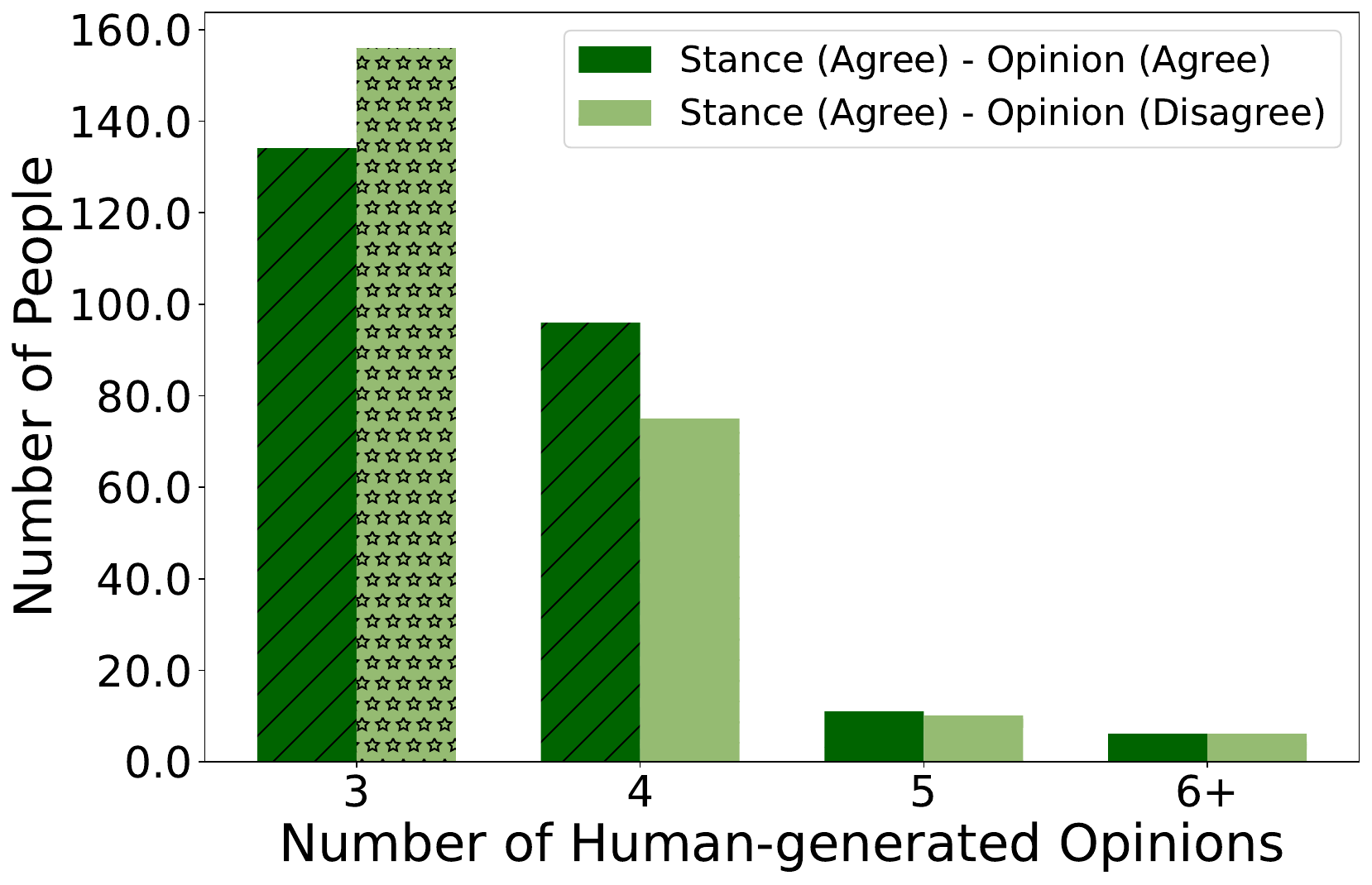}
        \caption{Human Stance: \textit{Agree}}
        \label{fig:human:task4:sc-agree}
    \end{subfigure}
    \begin{subfigure}[b]{0.45\textwidth}
        \centering
        \includegraphics[width=1.05\linewidth]{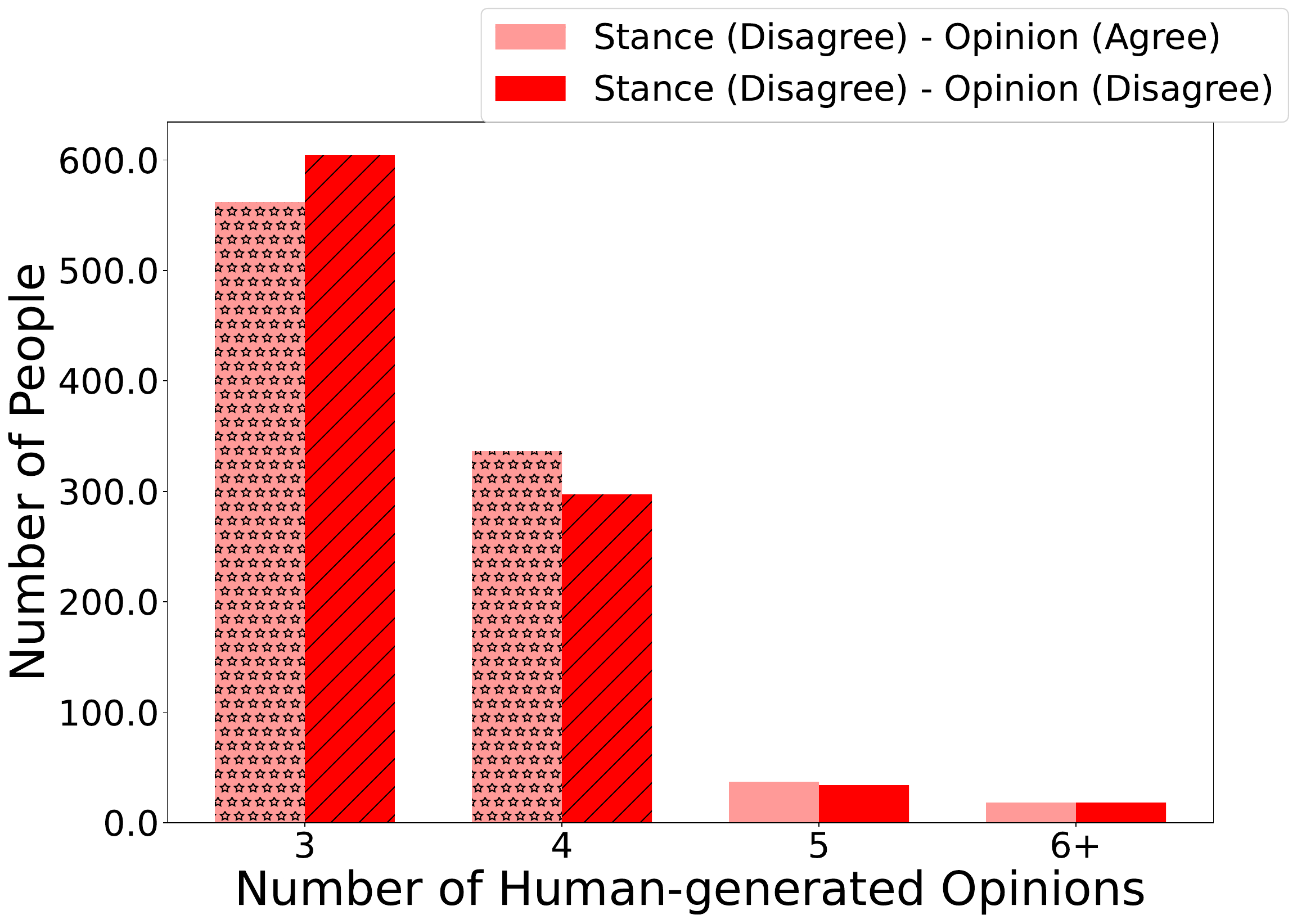}
       \caption{Human Stance: \textit{Disagree}}
        \label{fig:human:task4:sc-disagree}
    \end{subfigure}
    \label{fig:human:task4:sc}
     \begin{subfigure}{0.45\textwidth}
        \centering
        \includegraphics[width=\linewidth]{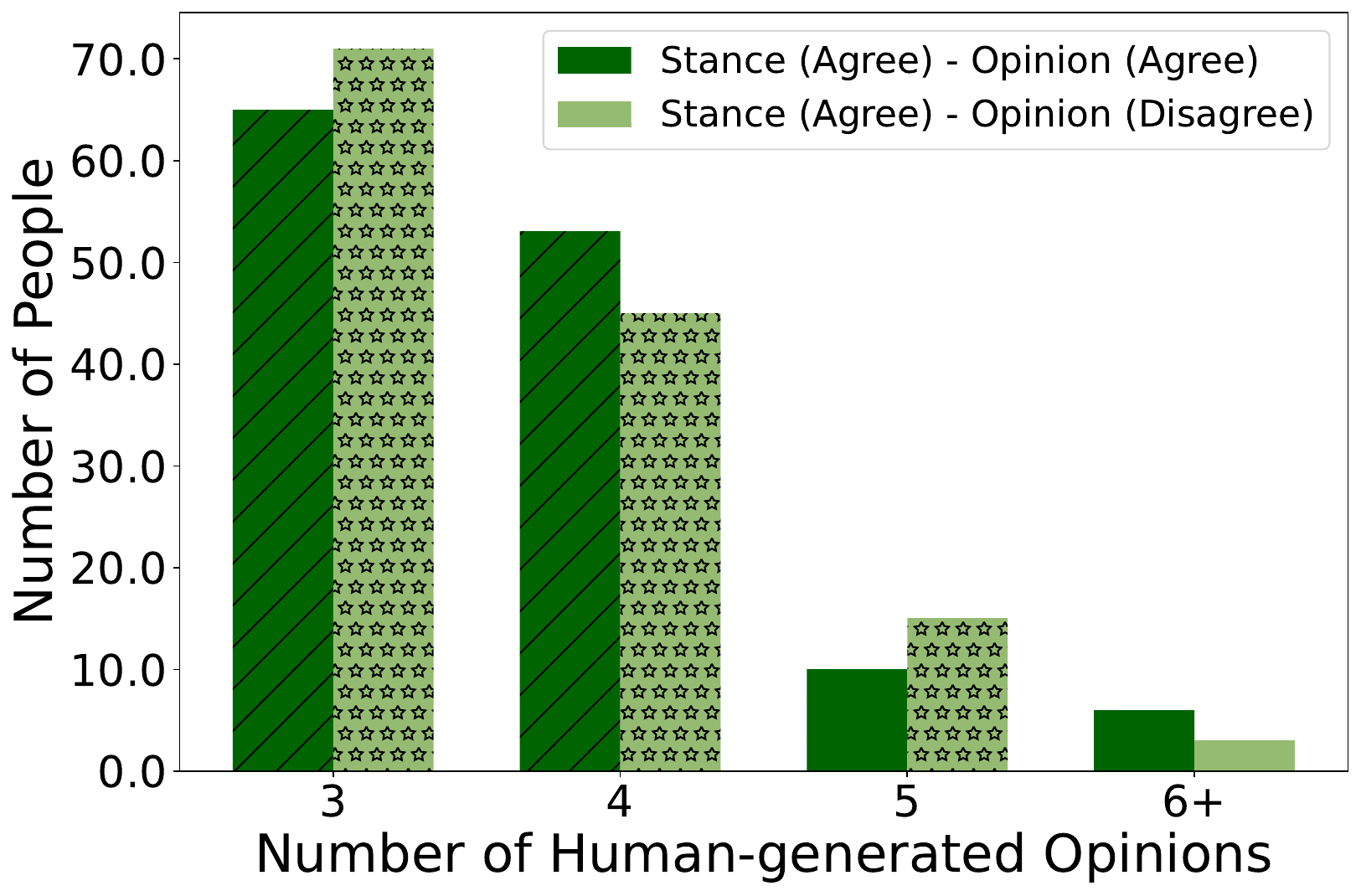}
        \caption{Human Stance: \textit{Agree}}
        \label{fig:human:task4:cmv-agree}
    \end{subfigure}
    \hfill 
    \begin{subfigure}[b]{0.45\textwidth}
        \centering
        \includegraphics[width=\linewidth]{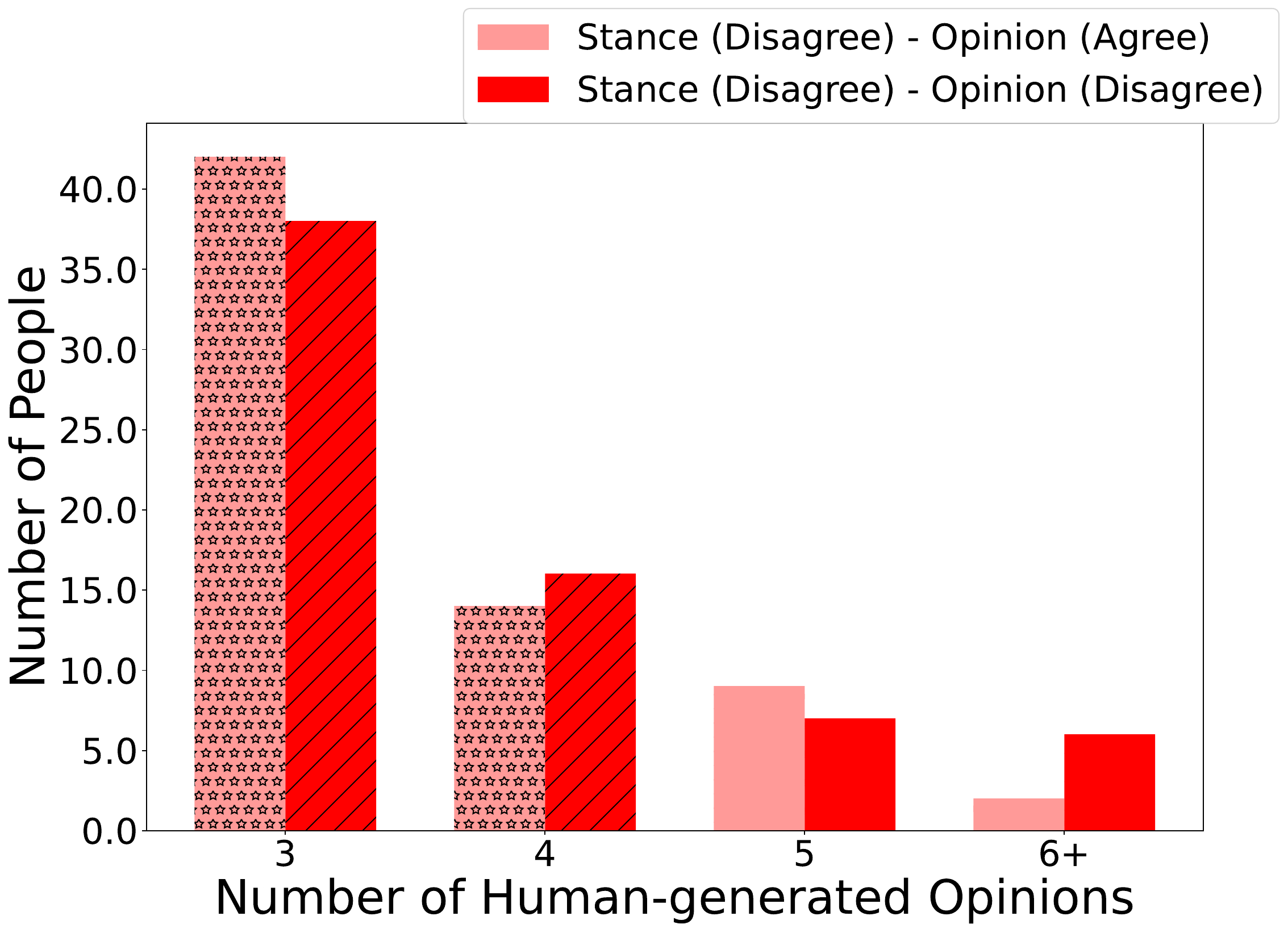}
       \caption{Human Stance: \textit{Disagree}}
        \label{fig:human:task4:cmv-disagree}
    \end{subfigure}
    \caption{The distribution of human-written opinions, separated by the human stance on given statements, in \textsc{Social-chem-101} (top) and \textsc{CMV} (bottom). Paler-colored bars represent instances where participants were asked to write opinions that opposed their personal stances on a statement. Each bar indicates the count of participants who provided the number of opinions corresponding to the bar's position on the x-axis.}
    \label{fig:entire-task4-appendix}
\end{figure*}

When participants from Section \ref{sec:human-vs-llm} are asked to provide opinions more than 3 about statements from \textsc{Social-Chem-101} and \textsc{CMV}, they tend to provide more opinions to the stance that they support, as shown in Figure \ref{fig:entire-task4-appendix}. This may indicate that humans are likely to be more conservative toward the stance against their preference. We suggest that humans' interaction with LLMs powered by our criteria-based prompting can also maximize diversity within the human perspectives, creating more inclusive environments within a society. Feasible future work could also be the evaluation of criteria-based prompting on human perspectives in an interactive setting. 

\subsection{Details about AMT Experiment Setups} \label{sec:appendix:human}

\paragraph{For Section \ref{sec:human-vs-llm},} we engaged three workers from the Amazon Mechanical Turk (AMT) platform, each of whom (1) lives in one of the five English-speaking countries (U.S., Canada, Australia, New Zealand, and the United Kingdom), (2) achieved a Human Intelligence Task (HIT) approval rate of 99\% or higher, as well as (3) the number of HITs approved greater than 10000 on the platform. Each HIT consists of five statements, and per statement each worker was supposed to provide at least 3 opinions that both agree and disagree with the statement, regardless of their personal stance on the statement. 

To acquire a pool of workers with better-quality responses, we manually reviewed every response from HITs once provided, filtering out the responses that fell short into the following types: (1) irrelevant to our statements; (2) not explaining your rationales behind your stance; or (3) saying that you just don't want to provide reasons to any stance. For example, if a participant answers like `N/A', `Nothing', `I don't agree/disagree', `Good', `Hello', etc. all of which are nonsense, these responses were not accepted.  We compensated those participants with \$2 USD payment for their participation. 

Tables \ref{table:appendix:demo-age}, \ref{table:appendix:demo-race}, and \ref{table:appendix:demo-ed} presents the demographic details of the participants of Section \ref{sec:human-vs-llm}, for both \textsc{Social-Chem-101} and \textsc{CMV} datasets.

\begin{table}[ht!]
\centering
\resizebox{\columnwidth}{!}{%
\begin{tabular}{c|c|c}
\toprule\toprule
Age group & \begin{tabular}[c]{@{}c@{}}\# Participants\\ (\textsc{Social-Chem-101})\end{tabular} & \begin{tabular}[c]{@{}c@{}}\# Participants\\ (\textsc{CMV})\end{tabular} \\ \midrule\midrule
17 - 24 & 10  & 7 \\ \hline
25 - 34 & 29 & 19 \\ \hline
35 - 44 & 10 & 7 \\ \hline
45 - 54 & 7  &  6\\ \hline
55 - 64  & 2  &  2\\ \hline
65 - 74 & 2  &  1\\ \bottomrule\bottomrule
\end{tabular}}
\caption{Demographic statistics of the participants of Section \ref{sec:human-vs-llm} - (1) Age Group. Most of the participants belong to the age group between 25 to 44.}
\label{table:appendix:demo-age}
\end{table}

\begin{table}[ht!]
\centering
\resizebox{\columnwidth}{!}{%
\begin{tabular}{c|c|c}
\toprule\toprule
Race group & \begin{tabular}[c]{@{}c@{}}\# Participants\\ (\textsc{Social-Chem-101})\end{tabular} & \begin{tabular}[c]{@{}c@{}}\# Participants\\ (\textsc{CMV})\end{tabular} \\ \midrule\midrule
White & 47  &  34\\ \hline
Asian & 1 & 1 \\ \hline
Latino & 1 & 0 \\ \hline
Black & 3 &  2\\ \hline
Native  & 7  & 4\\ \hline
Pacific  & 1  & 1 \\ \bottomrule\bottomrule
\end{tabular}}
\caption{Demographic statistics of the participants of Section \ref{sec:human-vs-llm} - (2) Race Group. Almost all of the participants identified themselves as White.}
\label{table:appendix:demo-race}
\end{table}

\begin{table}[ht!]
\centering
\resizebox{\columnwidth}{!}{%
\begin{tabular}{c|c|c}
\toprule\toprule
 \begin{tabular}[c]{@{}c@{}}Highest \\Education \\Level \end{tabular}& \begin{tabular}[c]{@{}c@{}}\# Participants\\ (\textsc{Social-Chem-101})\end{tabular} & \begin{tabular}[c]{@{}c@{}}\# Participants\\ (\textsc{CMV})\end{tabular} \\ \midrule\midrule
High School & 3  & 3 \\ \hline
Associate & 7 & 4 \\ \hline
Bachelor's & 45 &  33\\ \hline
Master's & 5  & 2 \\ \bottomrule\bottomrule
\end{tabular}}
\caption{Demographic statistics of the participants of Section \ref{sec:human-vs-llm} - (3) Highest Education Attainment. We observed most of the participants obtained bachelor's degrees.}
\label{table:appendix:demo-ed}
\end{table}

\paragraph{For Section \ref{sec:human:task1b}} We recruited three workers from the Amazon Mechanical Turk (AMT) platform, each of whom (1) lives in one of the five English-speaking countries (U.S., Canada, Australia, New Zealand, and the United Kingdom), (2) achieved a Human Intelligence Task (HIT) approval rate of 98\% or higher, as well as (3) the number of HITs approved greater than 10000 on the platform. Each HIT consists of five statements and a pairwise comparison set of A and B, where A and B are either criteria-prompting or free-form prompting outputs under anonymity. 

For each statement, workers were then asked to provide their own stance on that statement and choose between A and B in terms of which set possesses a more diverse perspective toward the statement. To acquire a pool of workers with better-quality responses, we ran a training session before the real task, where we also filtered out the workers whose responses did not make sense. Only the workers who showed above a threshold of our own were able to proceed to the real tasks. For their efforts, each participant received at least \$0.5 USD payment for completing the HIT regardless of the quality of their responses, considering the simplicity of the task and an anticipated time of completion to be less than 15 minutes. For those who showed a well-done performance that passed our threshold, we remunerated each of them with a bonus of at least \$2 (USD). 

First, we randomly sampled 50 statements from a dataset. For each statement, we created two  separate evaluation sessions, one for the 'agree' stance and the other for 'disagree.' In each session, we present the two sets of opinions that have that stance and are generated by GPT-3.5 in an one-shot setting: (1) Set A, generated with criteria-based prompting, and (2) Set B with free-form prompting. The workers were first asked to choose their own stance on the statement. Then, for each session, they select either set A or B, which they think includes a greater diversity of perspectives that show the same stance. 

After the response collection, we aggregated the majority of workers' preferences per statement under each stance session, and if the majority preference is the set generated by criteria-based prompting, we considered that criteria-prompting wins over the free-form prompting for the statement. Lastly, we computed the proportion of statements where criteria-based prompting wins, out of the 50 statements (denoted as ``win rate of criteria-based prompting'').

\subsection{Interface Design} \label{sec:appendix:interface}

\begin{figure*}
\centering
    \includegraphics[width=\linewidth]{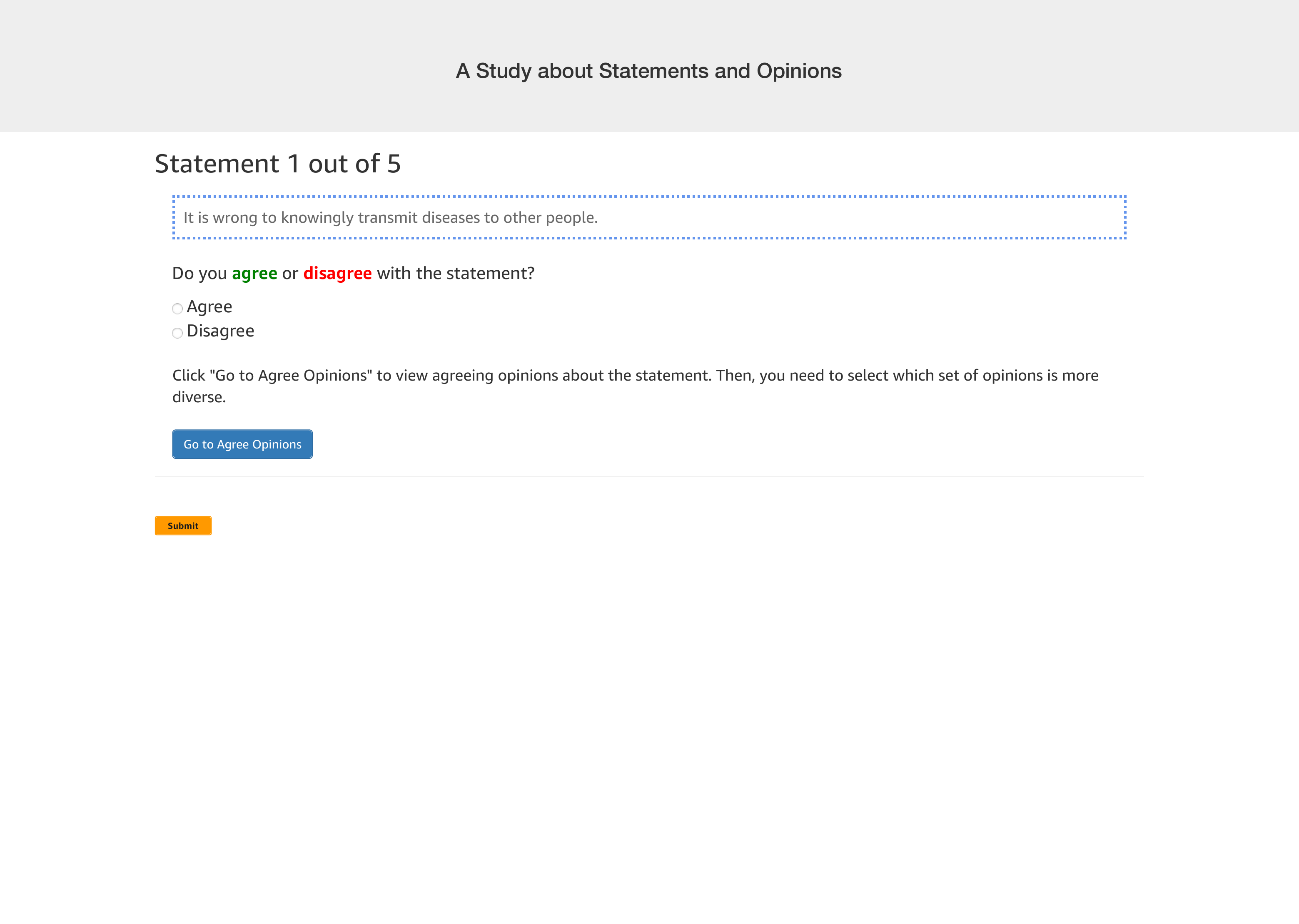}
    \includegraphics[width=\linewidth]{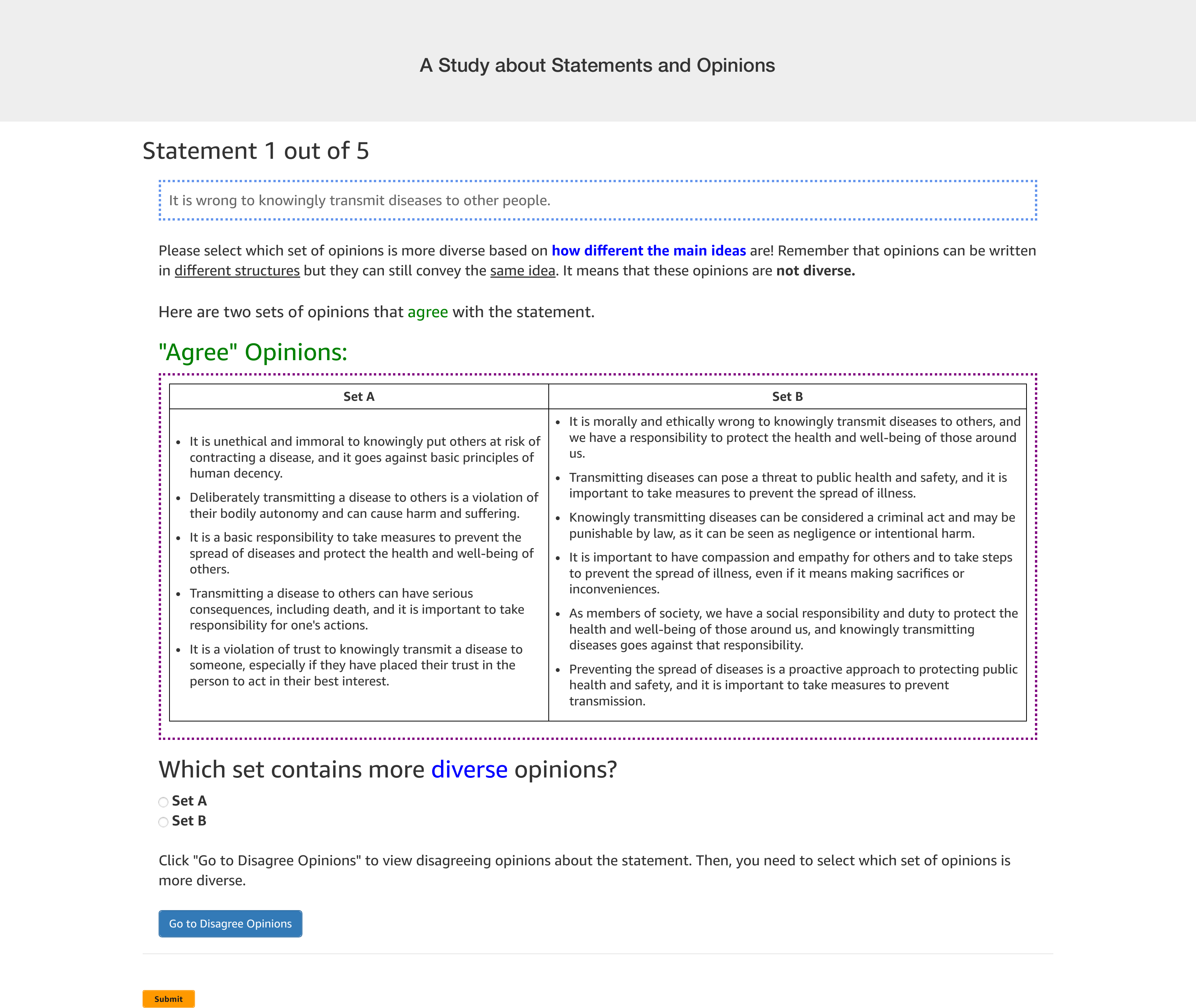}
        \caption{The AMT interface design for the human evaluation experiment in Section \ref{sec:human:task1b}.}
        \label{fig:appendix:task1b-interface}
\end{figure*}

\begin{figure*}[t]
    \centering
    \begin{subfigure}{\textwidth}
        \centering
        \includegraphics[width=0.6\linewidth]{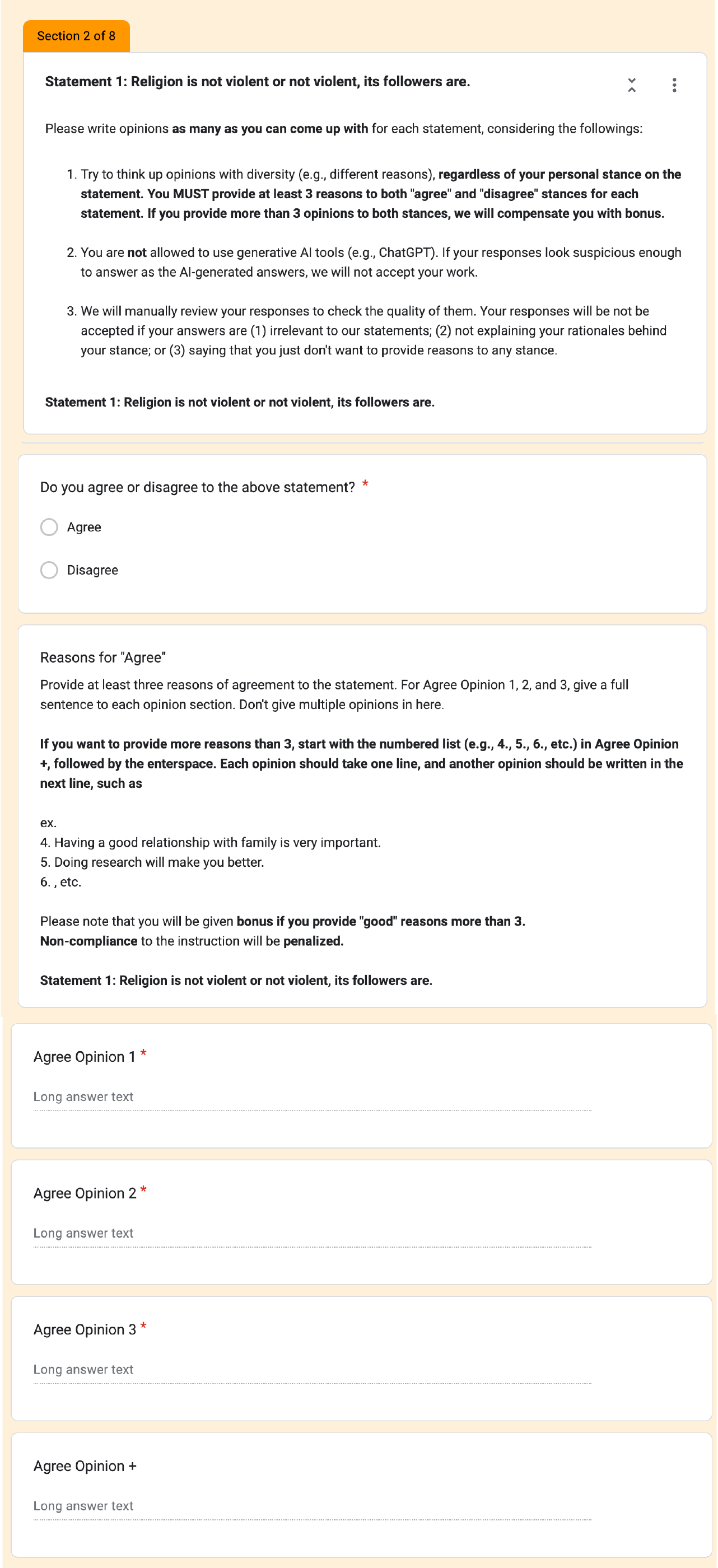}
    \end{subfigure}
        \caption{The AMT interface design for gathering human opinions in Section \ref{sec:human-vs-llm} - (1) The section for `Agree'}
        \label{fig:appendix:human-vs-llm-disagree}
\end{figure*}

We present the interface design templates for each of the AMT experiment setups (Section \ref{sec:human-vs-llm} and \ref{sec:human:task1b}) in Figures \ref{fig:appendix:task1b-interface} and \ref{fig:appendix:human-vs-llm-disagree}. The original prototype of Figure \ref{fig:appendix:task1b-interface} is referenced from \citet{hayati-etal-2021-bert}.

\begin{figure*}[t]
    \centering
    \begin{subfigure}{\textwidth}
        \centering
        \includegraphics[width=0.6\linewidth]{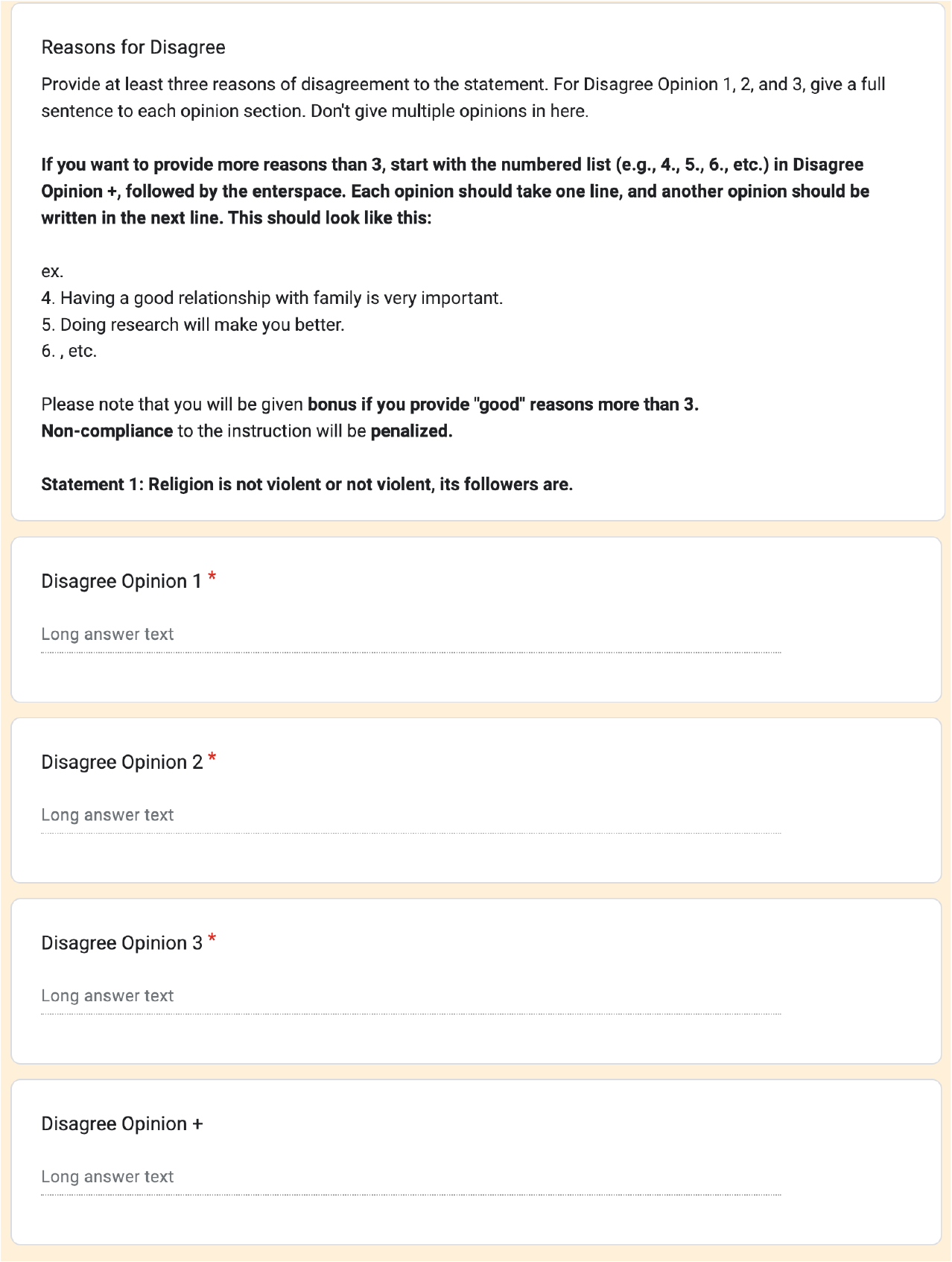}
    \end{subfigure}
        \caption{The AMT interface design for gathering human opinions in Section \ref{sec:human-vs-llm} - (2) The section for `Disagree'}
        \label{fig:appendix:human-vs-llm-agree}
    \end{figure*}

\end{document}